\documentclass[acmtog]{acmart}

\setcopyright{none}
\renewcommand\footnotetextcopyrightpermission[1]{}
\usepackage[shortlabels]{enumitem}
\usepackage{enumitem}

\usepackage{enumitem,lipsum}
\usepackage{rotating}
\usepackage{multirow}
\usepackage{wrapfig}

\usepackage{booktabs} 
\usepackage{overpic}
\usepackage[ruled]{algorithm2e} 

\SetAlFnt{\small}
\SetAlCapFnt{\small}
\SetAlCapNameFnt{\small}
\SetAlCapHSkip{0pt}

\begin{document}

\title{NeurCross: A Neural Approach to Computing Cross Fields for Quad Mesh Generation}

\author{Qiujie Dong}
    \orcid{0000-0001-6271-2546}
    \affiliation{%
    \institution{Shandong University}
    \city{Qingdao}
    \state{Shandong}
    \country{China}}
    \affiliation{%
    \institution{The University of Hong Kong}
    \city{Hong Kong}
    \country{China}}
    \affiliation{%
    \institution{TransGP}
    \city{Hong Kong}
    \country{China}}
    \email{qiujie.jay.dong@gmail.com}

\author{Huibiao Wen}
    \orcid{0009-0008-3589-7459}
    \affiliation{%
    \institution{Shandong University}
    \city{Qingdao}
    \state{Shandong}
    \country{China}}
    \email{ericvein@163.com}
    
\author{Rui Xu}
    \orcid{0000-0001-8273-1808}
    \affiliation{%
    \institution{The University of Hong Kong}
    \city{Hong Kong}
    \country{China}}
    \email{xrvitd@163.com}
    
\author{Shuangmin Chen}
    \orcid{0000-0002-0835-3316}
    \affiliation{%
    \institution{Qingdao University of Science and Technology}
    \city{Qingdao}
    \state{Shandong}
    \country{China}}
    \email{csmqq@163.com}

\author{Jiaran Zhou}
    \orcid{0000-0002-2943-2806}
    \affiliation{%
    \institution{Ocean University of China}
    \city{Qingdao}
    \state{Shandong}
    \country{China}}
    \email{zhoujiaran@ouc.edu.cn}
    
\author{Shiqing Xin}
\authornote{Corresponding author: Shiqing Xin.}         
    \orcid{0000-0001-8452-8723}
    \affiliation{%
    \institution{Shandong University}
    \city{Qingdao}
    \state{Shandong}
    \country{China}}
    \email{xinshiqing@sdu.edu.cn}
    
\author{Changhe Tu}
    \orcid{0000-0002-1231-3392}
    \affiliation{%
    \institution{Shandong University}
    \city{Qingdao}
    \state{Shandong}
    \country{China}}
    \email{chtu@sdu.edu.cn}

\author{Taku Komura}
    \orcid{0000-0002-2729-5860}
    \affiliation{%
    \institution{The University of Hong Kong}
    \city{Hong Kong}
    \country{China}}
    \email{taku@cs.hku.hk}

\author{Wenping Wang}
    \orcid{0000-0002-2284-3952}
    \affiliation{%
    \institution{Texas A\&M University}
    \state{Texas}
    \country{United States of America}}
    \email{wenping@tamu.edu}

\begin{abstract}
Quadrilateral mesh generation plays a crucial role in numerical simulations within Computer-Aided Design and Engineering (CAD/E). Producing high-quality quadrangulation typically requires satisfying four key criteria. First, the quadrilateral mesh should closely align with principal curvature directions. Second, singular points should be strategically placed and effectively minimized. Third, the mesh should accurately conform to sharp feature edges. Lastly, quadrangulation results should exhibit robustness against noise and minor geometric variations.
Existing methods generally involve first computing a regular cross field to represent quad element orientations across the surface, followed by extracting a quadrilateral mesh aligned closely with this cross field. A primary challenge with this approach is balancing the smoothness of the cross field with its alignment to pre-computed principal curvature directions, which are sensitive to small surface perturbations and often ill-defined in spherical or planar regions.

To tackle this challenge, we propose \textit{NeurCross}, a novel framework that simultaneously optimizes a cross field and a neural signed distance function (SDF), whose zero-level set serves as a proxy of the input shape.  Our joint optimization is guided by three factors: faithful approximation of the optimized SDF surface to the input surface, alignment between the cross field and the principal curvature field derived from the SDF surface, and smoothness of the cross field.
Acting as an intermediary, the neural SDF contributes in two essential ways. First, it provides an alternative, optimizable base surface exhibiting more regular principal curvature directions for guiding the cross field. Second, we leverage the Hessian matrix of the neural SDF to implicitly enforce cross field alignment with principal curvature directions, thus eliminating the need for explicit curvature extraction.
Extensive experiments demonstrate that NeurCross outperforms the state-of-the-art methods in terms of singular point placement, robustness against surface noise and surface undulations, and alignment with principal curvature directions and sharp feature curves.

\end{abstract}

%
%


%
%

\begin{teaserfigure}
    \centering
    \includegraphics[width=\textwidth]{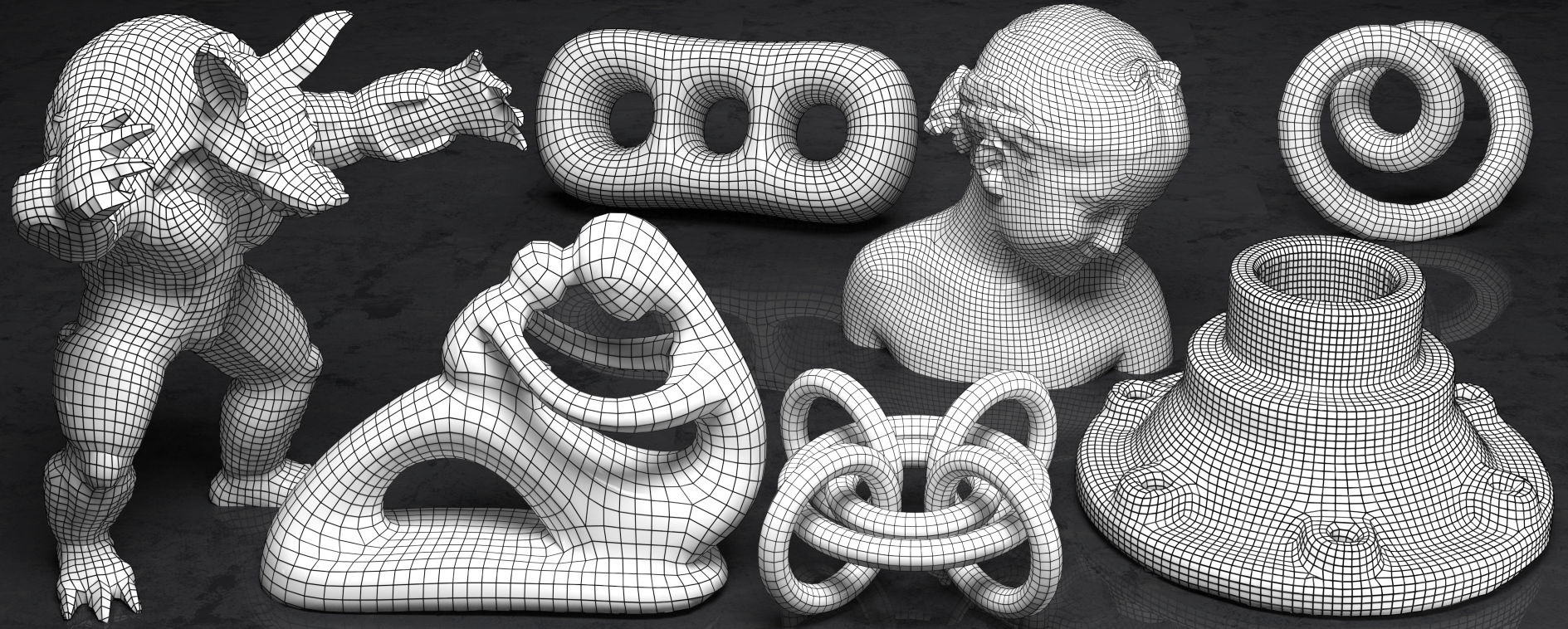}
    \vspace{-6mm}
    \caption{
Gallery of quad meshes generated with our NeurCros method.    
NeurCross excels in computing cross field for generating high-quality quad meshes. Its advantages include optimized singular point placement, insensitivity to surface noise and minor surface undulations, and faithful alignment with principal curvature directions and sharp feature curves.
    }
    \label{fig:teaser}
\end{teaserfigure}

\maketitle

\begin{figure*}[!t]
    \centering
    \begin{overpic}[width=\linewidth]{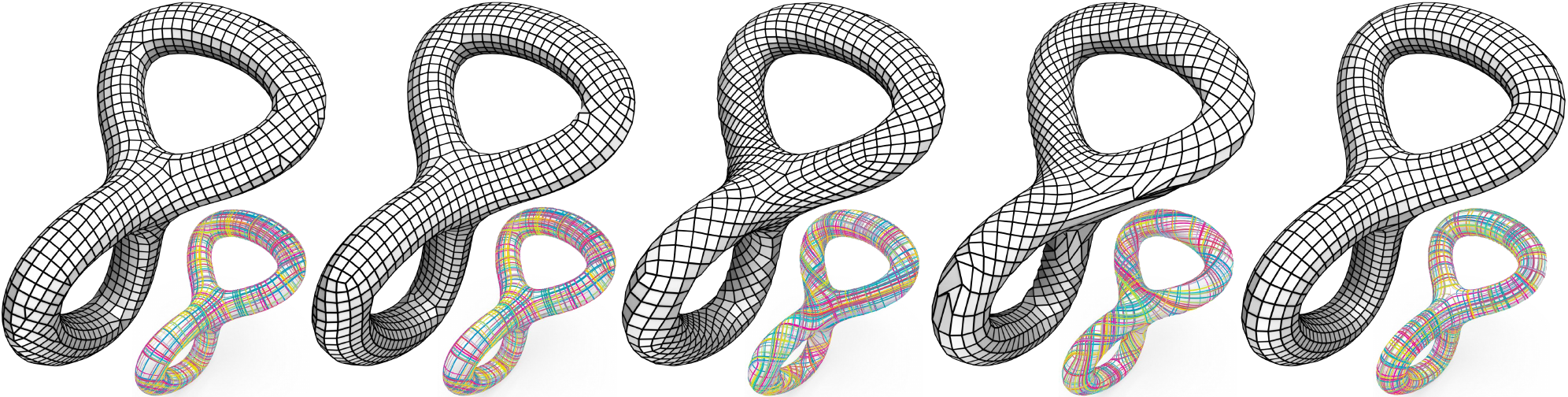}
        \put(7.5,-2){\textbf{IM}}
        \put(25,-2){\textbf{QuadriFlow}}
        \put(44,-2){\textbf{QuadWild}}
        \put(65,-2){\textbf{MIQ}}
        \put(82,-2){\textbf{NeurCross (Ours)}}
    \end{overpic}
    \vspace{-2mm}
    \caption{
  Existing approaches typically rely on principal curvature directions as input. However, due to the inherent instability of these directions, current methods often prioritize the smoothness of the cross field at the cost of alignment with the principal curvature directions. To address this limitation, our approach avoids explicitly extracting principal curvature directions. Instead, we assess whether the cross field at each point can function as eigenvectors of the shape operator.
    }
    \label{fig:double_tours}
\end{figure*}

\begin{figure*}[!t]
    \centering
    \begin{overpic}[width=\linewidth]{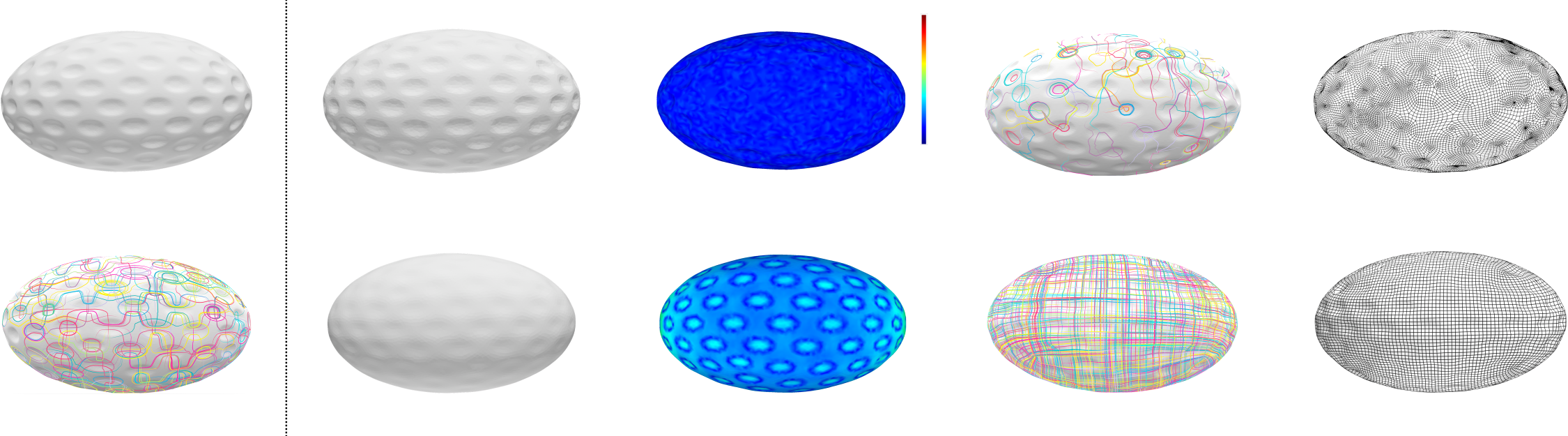}
        \put(0.5,26.8){\textbf{(a) Input}}
        \put(4.3,14.5){\textbf{GT surface}}
        \put(21,26.8){\textbf{(b) Two-step method}}
        \put(21,12.5){\textbf{(c) Our joint optimization}}
        \put(59.3,18.5){\small{\textbf{0}}}
        \put(59.3,26.2){\small{\textbf{0.001}}}
    
        \put(1.1,0.4){\textbf{GT curvature field}}
        \put(25.3,0.4){\textbf{SDF shape}}
        \put(45.5,0.4){\textbf{Fitting error}}
        \put(67,0.4){\textbf{Cross field}}
        \put(88,0.4){\textbf{Quad mesh}}
    \end{overpic}
    \vspace{-5mm}
    \caption{
    (a) The input mesh and its ground-truth principal curvature directions.
(b) Two-step optimization: by first precomputing an SDF that precisely fits the input shape, the subsequent optimization step still suffers from sensitivity to minor geometric variations, failing to yield the desired cross field.
(c) Joint optimization: by treating the SDF as a proxy for the input shape, simultaneous optimization of the SDF and the cross field allows the SDF to approximate the input shape while remaining robust to minor geometric variations, resulting in the desired cross field. We visualize the fitting errors between the SDF surface and the original surface using a color-coded scheme.}
    \label{fig:dimpled_ellipsoid}
    \vspace{-2mm}
\end{figure*}

\section{Introduction}
\label{sec:intro}
Quadrangulation is fundamental in both Computer-Aided Design (CAD) and Computer-Aided Engineering (CAE)~\cite{quadMeshsurvey2013, fieldDesign2016}, with significant applications in finite element analysis, isogeometric analysis, character animation, and physics simulations~\cite{IGM2013, MIQ2009, CBK15, Instant_Meshes2015, CVMJ}.

Existing approaches typically first compute a reliable cross field to represent quad element orientations across the surface, followed by extracting quad meshes aligned closely with the computed field \cite{MIQ2009, Instant_Meshes2015, QuadriFlow2018, IGM2013, Justin2020}. Most methods require principal curvature directions as input. However, computing a desired cross field from principal curvature directions entails meeting four key requirements: First, the quadrilateral mesh should align closely with principal curvature directions. Second, singular points should be strategically placed and minimized. Third, the mesh should conform accurately to sharp feature edges. Lastly, quadrangulation results should be robust against noise and minor surface variations. These challenges are particularly pronounced in geometrically or topologically complex shapes.

Fig.~\ref{fig:double_tours} shows quadrangulation results of some existing methods. As shown, for instance, QuadWild~\cite{quadwild2021} fails to align properly with principal curvature directions due to an overemphasis on the smoothness of the cross field. Although principal curvature directions provide useful geometric clues, precisely controlling their influence on the inferred cross field is difficult, especially in nearly spherical or planar regions, or on a surface with small undulations, where principal curvature directions become unstable.

We introduce an optimizable neural signed distance function (SDF) as the underlying shape representation to infer the desired cross field. The neural SDF serves as a proxy for the input shape, which often exhibits unstable principal curvature directions. Optimizing this SDF alongside the cross field provides a smooth approximation to the input shape, generating a regular principal curvature field to guide cross field generation. More specifically, our joint optimization is guided by three factors: faithful approximation of the optimized SDF surface to the input surface, alignment between the cross field and the principal curvature field derived from the SDF surface, and smoothness of the cross field. We integrate these requirements into a unified neural optimization framework, called {\em NeurCross}, that enables simultaneous optimization of the SDF and cross field. Fig.~\ref{fig:dimpled_ellipsoid} illustrates our method's success on a dimpled ellipsoid with irregular curvature directions, compared to a na\"{i}ve two-stage approach that first optimizes an SDF to properly fit the input shape and then uses the curvature field of this fixed SDF to guide the generation of the cross field. The key to the success of our method is its simultaneous optimization strategy that allows the cross field smoothness term to inform the optimal shape of the neural SDF as a proxy surface.

Additionally, a key advantage is that the SDF-based shape operator implicitly encodes principal curvature directions, enabling enforcement of alignment between principal curvature directions and the cross field by evaluating whether the cross at each point match well with the eigenvectors of the shape operator, bypassing the need for explicit extraction of principal curvature direction, a step susciptable to unstability in nearly spherical or planar regions.

\begin{figure*}[!t]
    \centering
    \begin{overpic}[width=\linewidth]{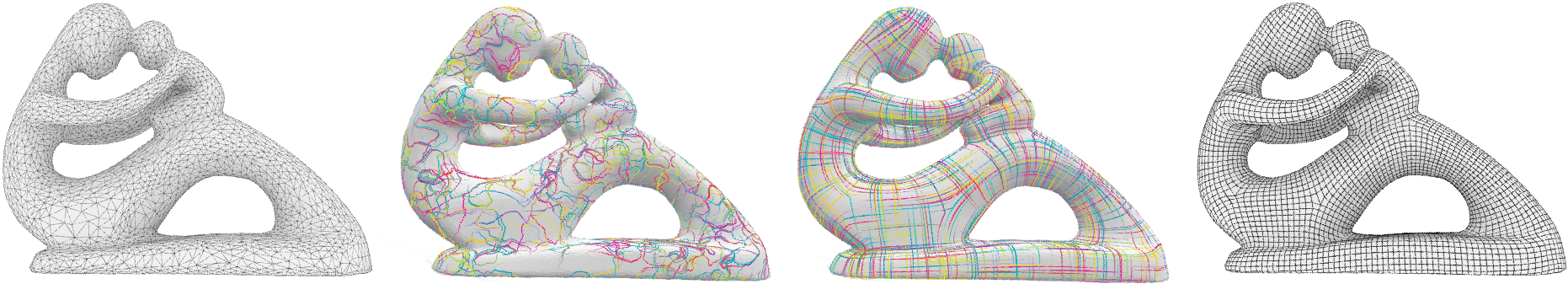}
    \put(5.3,-2){\textbf{(a) Triangle mesh}}
    \put(28,-2){\textbf{(b) Random initialization}}
    \put(54.3,-2){\textbf{(c) Resultant cross field}}
    \put(83,-2){\textbf{(d) Quad mesh}}
    \end{overpic}
    \vspace{-2mm}
    \caption{
    Given the input triangular surface in (a), starting with a randomly initialized cross field in (b),  our {\em NeurCross} method produces a smooth cross field in (c) that is well aligned with the principal curvature directions of the input surface. We use the global-seamless parametrization from libigl to obtain a parametrization aligned with the computed cross field, and then use libQEx to extract the final quad mesh in (d).
    }
    \label{fig:cross_field}
    \vspace{-2mm}
\end{figure*}

We implement NeurCross using a SIREN-based~\cite{SIREN} module for SDF fitting and a U-Net-based~\cite{UNet2015} module for cross field prediction. Over 10,000 iterations, both components are optimized simultaneously to satisfy quadrangulation criteria. Finally, we employ global-seamless parametrization from libigl~\cite{libigl2017} aligned with our cross field, followed by quad mesh extraction using libQEx~\cite{libQEX13}. Fig.~\ref{fig:cross_field} illustrates this process. Extensive experiments validate NeurCross's effectiveness, demonstrating improvements in singular point placement, robustness to noise and geometric variations, and approximation accuracy, as shown in the teaser figure.

Our contributions are summarized as follows:
\begin{enumerate}[-,nosep]
 \item We propose {\em NeurCross}, the first self-supervised neural network for learning cross fields.
 \item We implicitly enforce cross field alignment with principal curvature directions via an SDF-based shape operator, naturally addressing potential ambiguity.
 \item We leverage an optimizable neural SDF as an underlying representation to coordinate requirements, dynamically adjusting to minor surface variations.
\end{enumerate}

\section{Related Work}
This paper focuses on developing a neural representation of the cross field for quadrilateral mesh generation. In this section, we review two main categories of related work: quad mesh generation techniques and neural SDF representations.

\vspace{-2mm}
\subsection{Quad Mesh Generation}

Quadrilateral mesh generation has attracted significant attention in recent years. While some methods, such as Dual Marching Cubes (DMC)~\cite{DMC}, can directly extract quad facets without relying on direction fields, the resulting meshes often lack quality, particularly in aligning with principal directions. Most state-of-the-art approaches rely on a cross field~\cite{RVLL08, LJX10, Justin2021Frame} to guide the generation of high-quality quad meshes, as it ensures edge alignment and proper placement of irregular vertices. Typically, after computing a cross field, a parameterization step~\cite{MIQ2009, LZ14, CLW16, MPZ14} aligns gradients with the direction field and traces integer iso-lines across multiple charts~\cite{libQEX13}. Although several robust quadrangulation methods~\cite{OSCS99, RLS12, VZ01, GLLR11, DBG06, ZHLB10, LHJ14} operate independently of direction fields, they often fail to achieve global smoothness. Below, we review related works on direction fields.

A fundamental requirement for direction fields is to align edge directions with principal curvature directions~\cite{curvatureField2000, fieldDesign2016, KNP07, Instant_Meshes2015, MIQ2009, IGM2013, QuadriFlow2018, FBT18, LCBK19}. \citet{LKH08} proposed an iterative relaxation scheme that incrementally aligns mesh edges with principal directions, though it requires additional post-processing to refine results. \citet{Instant_Meshes2015} introduced a unified local smoothing operator that optimizes both edge orientations and vertex positions in the output quad mesh. QuadriFlow~\cite{QuadriFlow2018} improved upon Instant Meshes~\cite{Instant_Meshes2015} by introducing linear and quadratic constraints, reducing singularities but struggling to preserve the original shape. Several methods~\cite{IGM2013, Instant_Meshes2015, QuadriFlow2018} optimize parametrization while incorporating integer constraints, a challenging mixed-integer programming (MIP) problem~\cite{MIQ2009} that is computationally intensive. These methods typically aim to minimize distortion and reduce singularities~\cite{IGM2013, MZ13, LZ14, MPZ14}. To address the computational complexity of IGM~\cite{IGM2013} on complex meshes, \citet{ESCK16} proposed a framework using efficient decimation and coarse-to-fine mapping to improve interactive performance. \citet{DL2quadMesh2021} introduced a learning-based approach for predicting direction fields, demonstrating its potential for quad mesh generation. However, its reliance on domain-specific networks and canonical alignment limits its generalizability and robustness to non-rigid changes.

\subsection{Neural SDF}
The Signed Distance Function (SDF) is a widely used geometric representation in computer graphics, particularly for surface reconstruction. For example, radial basis functions (RBF)~\cite{carr2001reconstruction} approximate the SDF, enabling the extraction of the target surface as the zero-isosurface of the SDF.

SDFs have also been extensively employed in deep learning-based surface reconstruction, including supervised implicit surface reconstruction methods~\cite{DeepSDF, Points2surf, GalerkinNG} and self-supervised approaches~\cite{PCP, NeuralPull, yifan2020isopoints}. For instance, IGR~\cite{IGR} incorporates the Eikonal term to enforce implicit geometric regularization, providing an effective mechanism for surface reconstruction. SIREN~\cite{SIREN} demonstrates that periodic activation functions are well-suited for representing complex natural signals and their derivatives using implicit neural representations. DiGS~\cite{DiGS} integrates Laplacian energy as a soft constraint for the SDF, proving effective for reconstructing surfaces from unoriented point clouds. Neural-Singular-Hessian~\cite{HessianZX} ensures that the Hessian of the neural implicit function has a zero determinant for points near the surface, which is particularly useful for recovering details from unoriented point clouds. Additionally, \citet{Dong2024NeurCADRecon} proposed a zero Gaussian curvature constraint for reconstructing CAD-type surfaces from low-quality unoriented point clouds. All these methods leverage neural networks to approximate the SDF.

In this paper, SDFs play a central role in quad mesh generation, as the Hessian of the SDF fully encodes principal curvatures and their directions~\cite{HessianZX, Dong2024NeurCADRecon}.

\begin{figure*}[!t]
    \centering
    \begin{overpic}[width=0.98\linewidth]{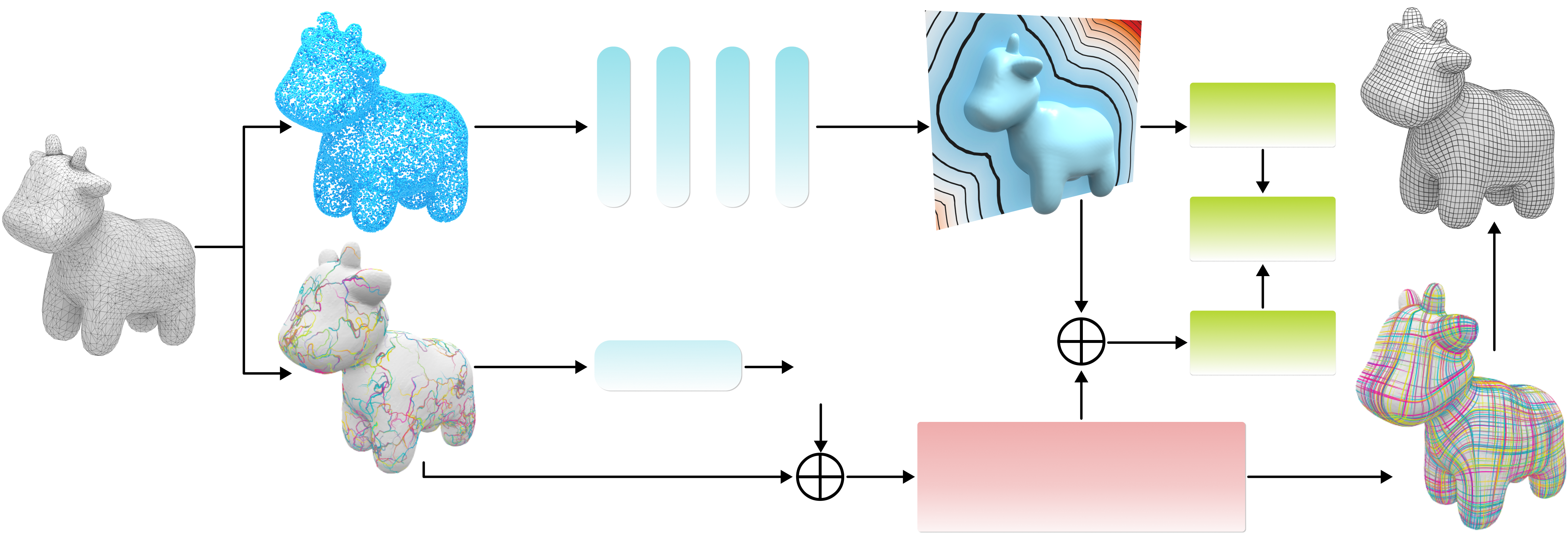}
    
    \put(32.5,26.5){$\mathcal{P}$}
    \put(30,11.5){$\mathcal{P}, \boldsymbol{n}, \boldsymbol{\mu}, \boldsymbol{\nu}$}
    \put(37.2,18.5){\textbf{SDF fitting module}}
    \put(40.8,10.1){MLP}
    \put(31,4.5){$\boldsymbol{\mu}, \boldsymbol{\nu}$}
    \put(50.7,9){{\fontsize{30}{0}\selectfont $\theta$}}
    \put(65.5,16){\fontsize{18}{0}\selectfont$\boldsymbol{H}$}
    \put(62,4.5){$\boldsymbol{\alpha}=\boldsymbol{\mu} \mathbf{cos}\theta + \boldsymbol{\nu} \mathbf{sin}\theta$}
    \put(62,1.5){$\boldsymbol{\beta}=\boldsymbol{\nu} \mathbf{cos}\theta - \boldsymbol{\mu} \mathbf{sin}\theta$}
    \put(76.5,12.3){$\mathcal{L}_\text{CrossField}$}
    \put(77.5,19){$\mathbf{min} \ \mathcal{L}$}
    \put(78,26.5){$\mathcal{L}_\text{SDF}$}  
    \end{overpic}
    \vspace{-2mm}
    \caption{
    Our self-supervised network pipeline for representing cross fields in quad mesh generation. All layers in the network are implemented as multi-layer perceptrons~(MLPs), with the SDF fitting module utilizing the SIREN~\cite{SIREN} architecture. The circled ``+'' symbol denotes a data-combining operation.
    }
    \label{fig:pipeline}
\end{figure*}

\section{Our Approach}

\subsection{Overview} 
The core idea of NeurCross is to leverage the optimizable neural Signed Distance Function (SDF) as an underlying representation to coordinate various requirements. On one hand, the adjustable SDF can effectively reduce sensitivity to minor surface variations. On the other hand, the SDF-based shape operator enables us to implicitly evaluate the difference between the principal curvature directions and the cross field. NeurCross consists of two core modules: a surface fitting module and an orientation prediction module, both centered around the SDF.
\begin{enumerate}
    \item \textbf{Surface Fitting Module:}
This module aims to represent the input triangular surface using a neural SDF. Its loss incorporates the Dirichlet condition~\cite{phase}, the Eikonal condition~\cite{IGR}, and the singular Hessian condition~\cite{RA2024alignHessian} to ensure high fidelity to the input surface geometry. Together, these constraints guarantee an accurate surface representation.

\item 
\textbf{Cross Field Prediction Module:}
This module is designed to represent the cross field while implicitly enforcing alignment with principal curvature directions and spatial smoothness. It employs a U-Net architecture~\cite{UNet2015} to predict a rotation angle for each triangular facet, yielding a geometry-aware cross field. Additionally, this module supports explicit alignment with geometric features.
 \end{enumerate}

These two modules are coordinated through a total loss function, which ensures simultaneous optimization of the SDF and the cross field during the training process. The interaction between the modules allows for dynamic updates to both the surface representation and the cross field, leading to improved accuracy and robustness.
The overall network architecture is illustrated in Fig.~\ref{fig:pipeline}.

\paragraph{Total Loss}
Our total loss is defined as follows.
\begin{small}
    \begin{equation}
\label{eq:our_loss}
    \mathcal{L} = \underbrace{\lambda_\text{E} \mathcal{L}_\text{E} + \lambda_\text{DM} \mathcal{L}_\text{DM} + \lambda_\text{DNM} \mathcal{L}_\text{DNM} + \tau\lambda_\text{AN} \mathcal{L}_\text{AN}}_{\text{SDF}} + \underbrace{\lambda_\text{AP} \mathcal{L}_\text{AP} + \lambda_\text{S} \mathcal{L}_\text{S}}_{\text{Cross Field}},
\end{equation}
\end{small}
where~$\tau$ is the annealing factor~\cite{HessianZX, RA2024alignHessian, Dong2024NeurCADRecon}.
The individual terms, along with their corresponding weights, will be detailed in the following subsections.

\subsection{SDF Fitting}
\label{sec:sdf}
Let $\Theta$ denote the parameters of a neural SDF $f(\boldsymbol{x}; \Theta): \mathbb{R}^3 \to \mathbb{R}$, where $f = 0$ approximates the input triangular surface. We begin by sampling the centroid of each triangle, forming a point set \(\mathcal{P}\), where each point is associated with a normal vector. In the following, we define a loss term to enforce alignment with the predefined surface normals, while leaving further details to Sec.~\ref{sec:imple_details}.

\paragraph{SDF Based Shape Operator}
The shape operator of a surface measures the rate of change of the unit normal in any direction, thereby describing how the shape changes in that direction. In differential geometry, it is very common to assume the surface has a parametric form, such that the shape operator defines a quadratic form on the tangent space, and the eigenvectors of the shape operator correspond exactly to the principal directions. In fact, the Hessian matrix of the SDF is closely related to the shape operator. For a point \(\boldsymbol{p}\) on the base surface, the Hessian matrix \(\boldsymbol{H}_{\boldsymbol{p}}\) of the SDF has an eigenvalue of 0, with its corresponding eigenvector being the normal vector \(\boldsymbol{n}_{\boldsymbol{p}}\)~\cite{HessianZX, Dong2024NeurCADRecon, RA2024alignHessian}. Simultaneously, the other two eigenvectors of \(\boldsymbol{H}_{\boldsymbol{p}}\) correspond to the two principal curvature directions.

\paragraph{Alignment with Predefined Surface Normals}
Since for a point \(\boldsymbol{p}\) sufficiently close to the base surface, the eigenvector corresponding to the zero eigenvalue of the Hessian matrix \(\boldsymbol{H}_{\boldsymbol{p}}\) of the SDF aligns with the normal vector \(\boldsymbol{n}_{\boldsymbol{p}}\) at \(\boldsymbol{p}\).
Given that the normal direction of \(\boldsymbol{p} \in \mathcal{P}\) can be directly obtained from the input triangle mesh,
we require the neural SDF to align with the predefined surface normal $\boldsymbol{n}_{\boldsymbol{p}}$ as follows:
\begin{equation}
\boldsymbol{H}_{\boldsymbol{p}} \cdot \boldsymbol{n}_{\boldsymbol{p}} = \boldsymbol{0}.
\end{equation}
 The overall alignment with predefined surface normals can be quantified as:
\begin{equation}
\mathcal{L}_\text{AN} = \frac{1}{|\mathcal{P}|} \int_{\mathcal{P}} \big| \boldsymbol{H}_{\boldsymbol{p}} \cdot \boldsymbol{n}_{\boldsymbol{p}} \big| \, \text{d}\boldsymbol{p}.
\end{equation}

\subsection{Cross Field Prediction}
\label{sec:field}

\paragraph{Local Coordinate System}
\label{sec:local_coor}
\begin{wrapfigure}{r}{3.5cm}
  \hspace*{-4mm}
  \centerline{
  \includegraphics[width=45mm]{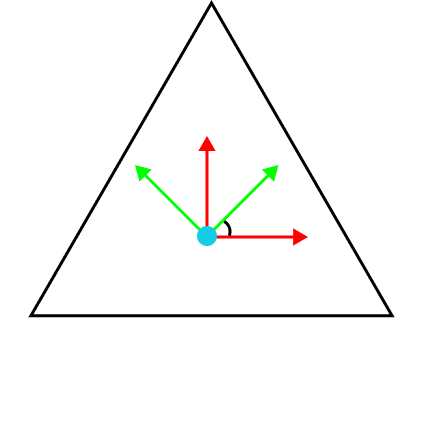}

  \put(-94,69){$\boldsymbol{\beta}_{\boldsymbol{p}}$}
  \put(-59,79){$\boldsymbol{\alpha}_{\boldsymbol{p}}$}

  \put(-56,60){$\theta_{\boldsymbol{p}}$}
  \put(-73,50){${\boldsymbol{p}}$}
  
  \put(-77,86){$\boldsymbol{\nu}_{\boldsymbol{p}}$}
  \put(-40,62){$\boldsymbol{\mu}_{\boldsymbol{p}}$}
  }
  \vspace*{-15mm}
\end{wrapfigure}
Recall that each point \(\boldsymbol{p} \in \mathcal{P}\) corresponds to the centroid of a triangular face. The task of computing the cross field involves inferring a pair of orthogonal vectors, \((\boldsymbol{\alpha}_{\boldsymbol{p}}, \boldsymbol{\beta}_{\boldsymbol{p}})\), that align as closely as possible with the principal curvature directions. To achieve this, we assume that each triangle has a pre-defined coordinate system with two axes, \(\boldsymbol{\mu}_{\boldsymbol{p}}\) and \(\boldsymbol{\nu}_{\boldsymbol{p}}\), which are mutually orthogonal unit vectors satisfying \(\boldsymbol{\mu}_{\boldsymbol{p}} \times \boldsymbol{\nu}_{\boldsymbol{p}} = \boldsymbol{n}_{\boldsymbol{p}}\). 
We introduce a rotation angle \(\theta_{\boldsymbol{p}}\) to represent \(\boldsymbol{\alpha}_{\boldsymbol{p}}\) and \(\boldsymbol{\beta}_{\boldsymbol{p}}\) as follows:
\begin{equation}
\left\{
    \begin{array}{l}
    \boldsymbol{\alpha}_{\boldsymbol{p}} = \boldsymbol{\mu}_{\boldsymbol{p}} \cos\theta_{\boldsymbol{p}} + \boldsymbol{\nu}_{\boldsymbol{p}} \sin\theta_{\boldsymbol{p}}, \\
    \boldsymbol{\beta}_{\boldsymbol{p}} = \boldsymbol{\nu}_{\boldsymbol{p}} \cos\theta_{\boldsymbol{p}} - \boldsymbol{\mu}_{\boldsymbol{p}} \sin\theta_{\boldsymbol{p}}.
    \end{array}
\right.
\end{equation}
See the inset figure for an illustration. Notably, \(\boldsymbol{\alpha}_{\boldsymbol{p}}\) and \(\boldsymbol{\beta}_{\boldsymbol{p}}\) are naturally mutually orthogonal unit vectors. As a result, the optimization of the cross field reduces to computing the rotation angle \(\theta_{\boldsymbol{p}}\) for each triangle.

\paragraph{Implicit Alignment with Principal Directions}
An explicit approach to implementing alignment with principal directions involves comparing the cross field with pre-extracted principal directions. However, most existing methods for extracting principal directions heavily rely on local shape variations, which can lead to instability, particularly when the local geometry is approximately planar or spherical. To address this limitation, we adopt an implicit alignment strategy by evaluating the compatibility between the cross field and the shape operator.

To align the cross field with the principal directions, we encourage \(\boldsymbol{\alpha}_{\boldsymbol{p}}\) and \(\boldsymbol{\beta}_{\boldsymbol{p}}\) to coincide with two of the eigenvectors of \(\boldsymbol{H}_{\boldsymbol{p}}\). 
To enforce collinearity between \(\boldsymbol{H}_{\boldsymbol{p}} \boldsymbol{\alpha}_{\boldsymbol{p}}\) and \(\boldsymbol{\alpha}_{\boldsymbol{p}}\), we impose the following condition:
\begin{equation}
\boldsymbol{H}_{\boldsymbol{p}} \boldsymbol{\alpha}_{\boldsymbol{p}} \times \boldsymbol{\alpha}_{\boldsymbol{p}} = \boldsymbol{0}.
\end{equation}
Similarly, we require:
\begin{equation}
\boldsymbol{H}_{\boldsymbol{p}} \boldsymbol{\beta}_{\boldsymbol{p}} \times \boldsymbol{\beta}_{\boldsymbol{p}} = \boldsymbol{0}.
\end{equation}
We define the loss term to measure alignment with the principal directions as follows:
\begin{equation}
    \mathcal{L}_\text{AP}^{(1)} = \frac{1}{|\mathcal{P}|}\int_{\mathcal{P}}{\big\vert \boldsymbol{H}_{\boldsymbol{p}} \boldsymbol{\alpha}_{\boldsymbol{p}} \times \boldsymbol{\alpha}_{\boldsymbol{p}} \big\vert + \big\vert \boldsymbol{H}_{\boldsymbol{p}} \boldsymbol{\beta}_{\boldsymbol{p}} \times \boldsymbol{\beta}_{\boldsymbol{p}} \big\vert}\text{d}\boldsymbol{p}.
\end{equation}

\paragraph{Smoothness of the Cross Field}
Consider two pairs of orthogonal unit vectors in a plane, denoted as $(\boldsymbol{\alpha}_1,\boldsymbol{\beta}_1)$ and $(\boldsymbol{\alpha}_2,\boldsymbol{\beta}_2)$. We say that the pair $(\boldsymbol{\alpha}_1,\boldsymbol{\beta}_1)$ aligns with $(\boldsymbol{\alpha}_2,\boldsymbol{\beta}_2)$ if either $\boldsymbol{\alpha}_1$ and $\boldsymbol{\alpha}_2$ are colinear, or $\boldsymbol{\alpha}_1$ and $\boldsymbol{\beta}_2$ are colinear.

Based on the above definition, it can be proved that
$(\boldsymbol{\alpha}_1,\boldsymbol{\beta}_1)$ aligns with $(\boldsymbol{\alpha}_2,\boldsymbol{\beta}_2)$
if and only if
\begin{equation}
    \big\vert\boldsymbol{\alpha}_1\cdot \boldsymbol{\alpha}_2\big\vert + \big\vert\boldsymbol{\alpha}_1\cdot \boldsymbol{\beta}_2\big\vert +  \big\vert\boldsymbol{\beta}_1\cdot \boldsymbol{\alpha}_2\big\vert + \big\vert\boldsymbol{\beta}_1\cdot \boldsymbol{\beta}_2\big\vert
\end{equation}
achieves the minimum.
We explain the correctness as follows.
Without loss of generality, 
we assume that $\boldsymbol{\alpha}_1=(1,0)$ and $\boldsymbol{\beta}_1=(0,1)$.
By denoting $\boldsymbol{\alpha}_2$ as $(\cos\theta,\sin\theta)$, 
the above sum simplifies to 
\begin{equation}
    2(|\cos\theta|+|\sin\theta|).
\end{equation}
\begin{wrapfigure}{r}{4cm}
\vspace{-5mm}
  \hspace*{-4.5mm}
  \centerline{
  \includegraphics[width=40mm]{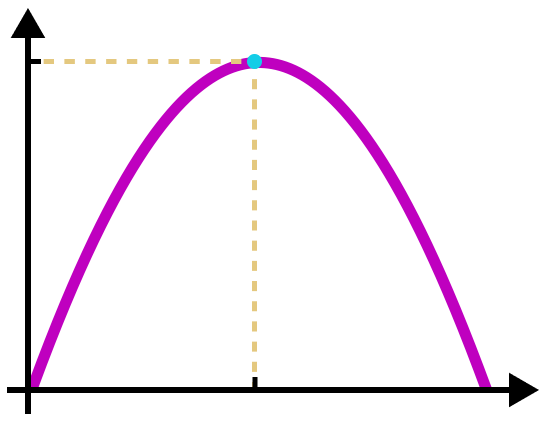}
		\put(-125,73){\textbf{$2\sqrt{2}$}}
		\put(-120,5){\textbf{2}}
        \put(-108,-4.0){\textbf{0}}
		\put(-66,-4.0){\textbf{$\pi/4$}}
		\put(-19,-4.0){\textbf{$\pi/2$}}
  }
  \vspace*{-4mm}
\end{wrapfigure}
In the inset figure, we illustrate how the function value of $2(|\cos\theta|+|\sin\theta|)$ varies with $\theta$. It can be observed that the minimum is achieved at $\theta = k\frac{\pi}{2}$, while the maximum occurs at $\theta = k\frac{\pi}{2} + \frac{\pi}{4}$. Therefore, it can be concluded that only when the sum reaches the minimum value of 2, one cross aligns with another ($\theta = k\frac{\pi}{2}$).

\begin{wrapfigure}{r}{3.5cm}
\vspace{-4mm}
  \hspace*{-9mm}
  \centerline{
  \includegraphics[width=35mm]{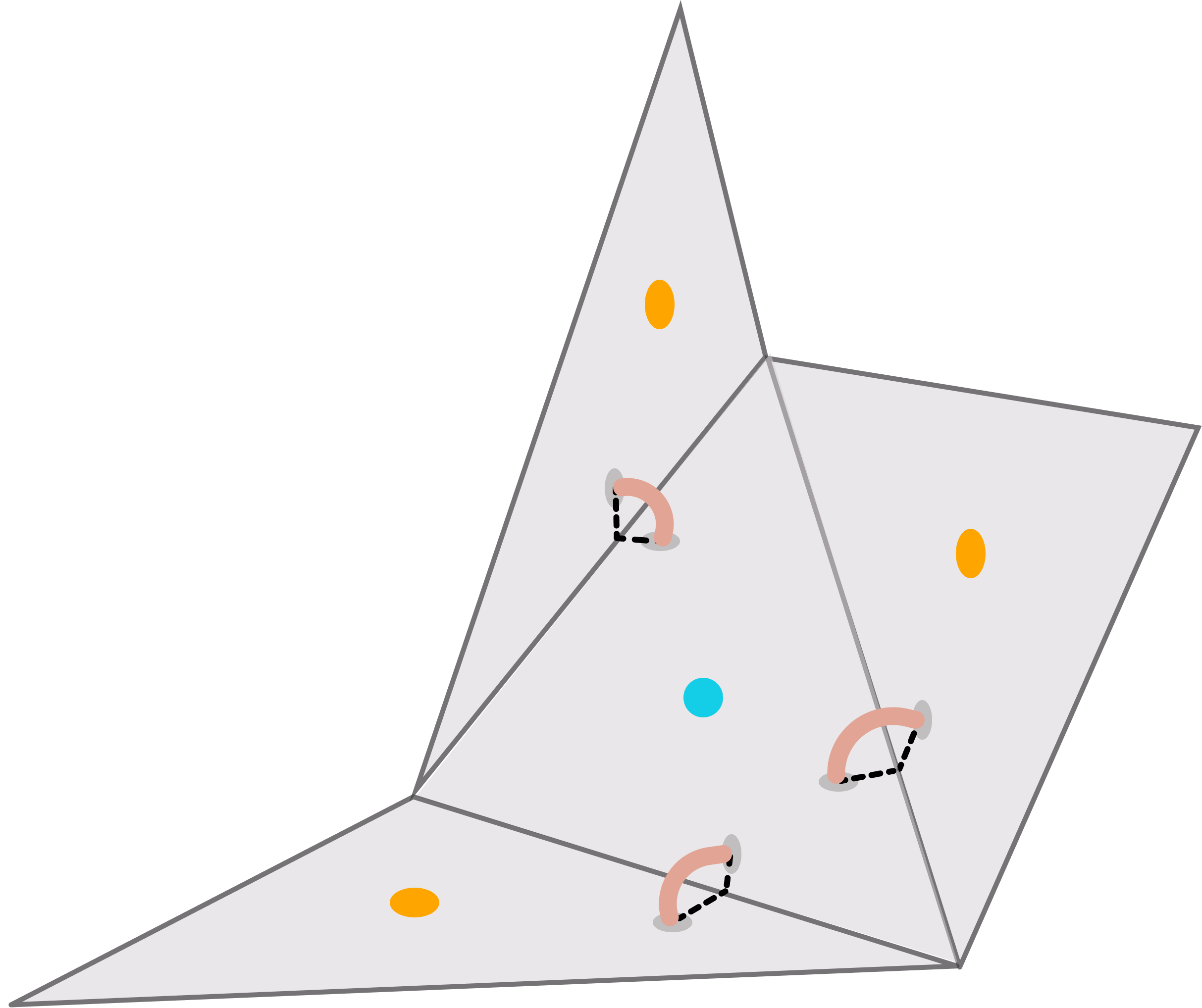}
		\put(-50,23){$\boldsymbol{p}$}
		\put(-75,8){$\boldsymbol{q}_1$}
        \put(-49,63){$\boldsymbol{q}_2$}
		\put(-18,40){$\boldsymbol{q}_3$}
		\put(-37,10){$\boldsymbol{\varphi}_1$}
        \put(-49,33){$\boldsymbol{\varphi}_2$}
        \put(-23.5,18){$\boldsymbol{\varphi}_3$}
  }
  \vspace*{-3mm}
\end{wrapfigure}
We denote the three neighboring points of $\boldsymbol{p}$ as $\boldsymbol{q}_1, \boldsymbol{q}_2, \boldsymbol{q}_3$. Since each $\boldsymbol{q}_i$ lies on a neighboring face, a rotation around the common edge is necessary before aligning the directions between $\boldsymbol{p}$ and $\boldsymbol{q}_i$. 
We define \(\{\boldsymbol{R}_i \mid i = 1, 2, 3\}\) as the rotation matrices associated with dihedral angles \(\{\boldsymbol{\varphi}_i \mid i = 1, 2, 3\}\) and shared edges, which can be precomputed.
To this end, the smoothness loss can be written as
\begin{equation}
\label{equ:coherence_term}
\begin{split}
    \mathcal{L}_\text{S} = \frac{1}{3|\mathcal{P}|}\int_{\mathcal{P}}\sum_{i=1}^3 & \left(\big\vert \boldsymbol{\alpha}_{\boldsymbol{p}}\cdot \boldsymbol{R}_i \boldsymbol{\alpha}_{\boldsymbol{q}_i} \big\vert + \big\vert \boldsymbol{\alpha}_{\boldsymbol{p}}\cdot \boldsymbol{R}_i \boldsymbol{\beta}_{\boldsymbol{q}_i} \big\vert \right. \\ & \left.  + \big\vert \boldsymbol{\beta}_{\boldsymbol{p}} \cdot \boldsymbol{R}_i\boldsymbol{\alpha}_{\boldsymbol{q}_i} \big\vert + \big\vert \boldsymbol{\beta}_{\boldsymbol{p}} \cdot \boldsymbol{R}_i\boldsymbol{\beta}_{\boldsymbol{q}_i} \big\vert-2 \right)\text{d}\boldsymbol{p}.
\end{split}
\end{equation}

\begin{figure}
    \centering
    \graphicspath{{figures/}}
    \begin{overpic}
		[width=0.95\linewidth]{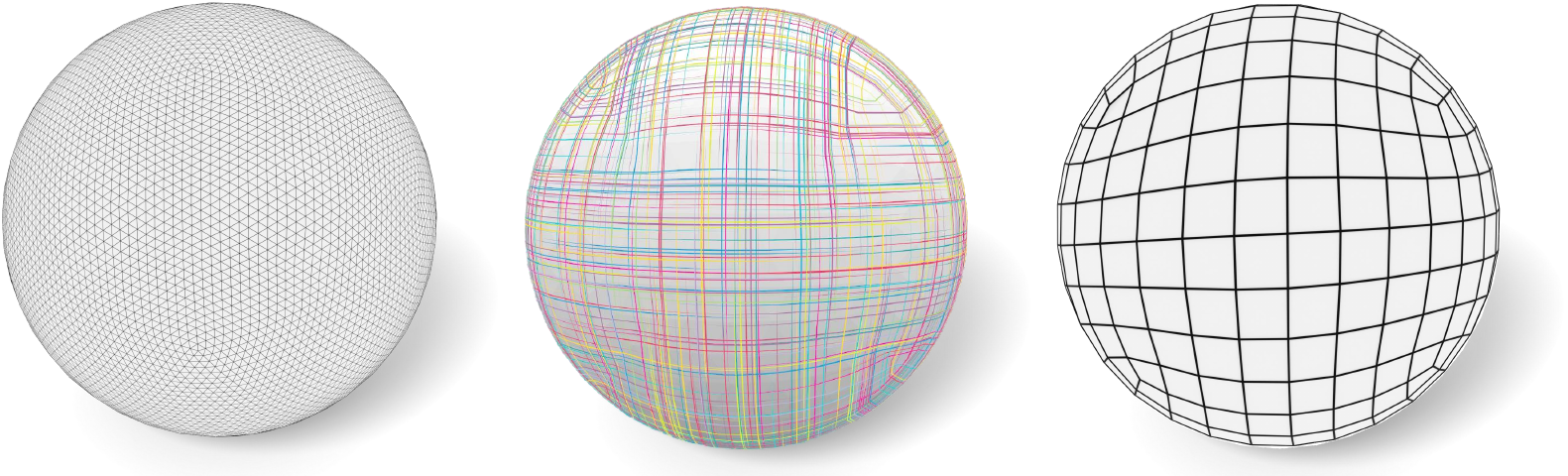}
        \put(1.2, -4.5){\textbf{(a) Input mesh}}
        \put(34.8, -4.5){\textbf{(b) Cross field}}
        \put(69, -4.5){\textbf{(c) Quad mesh}}
	\end{overpic}
   \vspace{2mm}
 \caption{The smoothness constraint of the cross field better controls the distribution of singularity points.
From left to right: (a) an input triangular mesh; (b) a cross field computed with NeurCross; (c) the quad mesh extracted from the cross field.
 }
 \label{fig:sphere}
 \vspace{0mm}
\end{figure}

{\bf Remark:} 
Consider a spherical surface, as shown in Fig.~\ref{fig:sphere}. At each point on such a surface, the principal curvature directions are not unique. In this case, the smoothness constraint of the cross field plays a crucial role in better controlling the distribution of singularity points. It is worth noting that our implicit principal curvature alignment is simultaneously satisfied.
Moreover, Fig.~\ref{fig:sphere} highlights the importance of effective cross field smoothing, which has also been addressed in prior work.
\citet{PowerFields_Felix2013} and~\citet{PolyVectors_Diamanti2014} introduced convex smoothness energies to smooth N-RoSy fields. \citet{PowerFields_Felix2013}’s method achieves global optimality but requires a nonlinear transformation, which can cause extra singularities and distortion. \citet{Instant_Meshes2015} uses an extrinsic energy to align with surface features, but its strong reliance on local information often leads to suboptimal results and unwanted singularities.
In contrast, our NeurCross smooths the cross field using Equ.~\ref{equ:coherence_term}, introducing singularities only in areas with high curvature variation. A detailed comparison is provided in Section~\ref{sec:more_comparison}.

\paragraph{Sharp Feature Alignment}
\label{sec:sharp_feature}
As pointed out in~\cite{quadwild2021}, in the context of quad-meshing, it is important to incorporate feature lines of the input shape, such as crease angles in CAD models, when generating quadrangulation outcomes. However, reconciling this
\begin{wrapfigure}{r}{4.5cm}
\vspace{-2mm}
  \hspace*{-4.5mm}
  \centerline{
  \includegraphics[width=50mm]{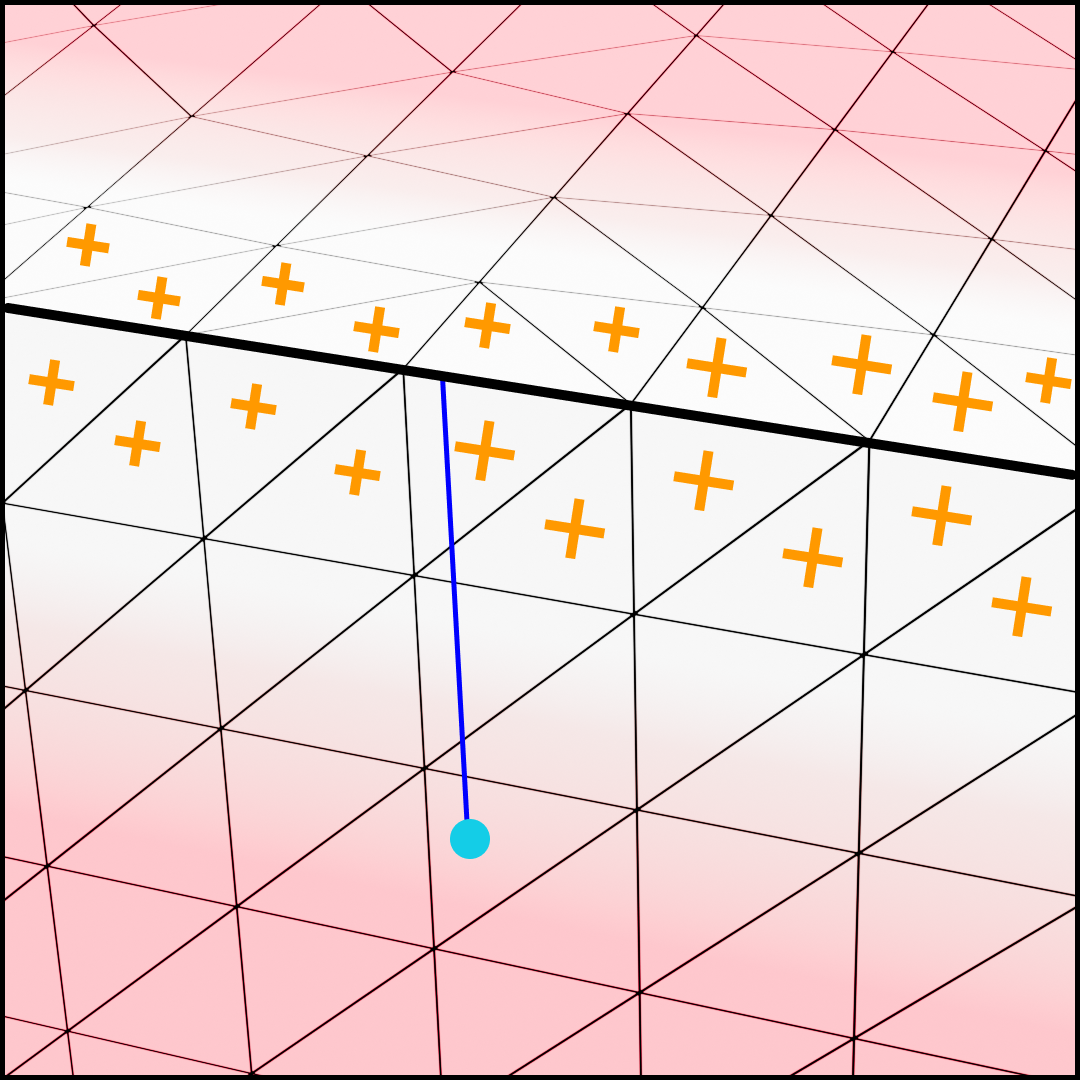}
	\put(-93,32){$\boldsymbol{p}$}
    \put(-75,93){$\boldsymbol{FL}$}
    \put(-75,50){$d_g(\boldsymbol{p}, FL)$}
  }
  \vspace*{-3mm}
\end{wrapfigure}
\hspace{-1.7mm} requirement with all ot- her objectives is challenging. Generally, the influence of feature lines diminishes with increasing distance.

\begin{figure}[t]
    \centering
    \graphicspath{{figures/}}
    \begin{overpic}
		[width=0.95\linewidth]{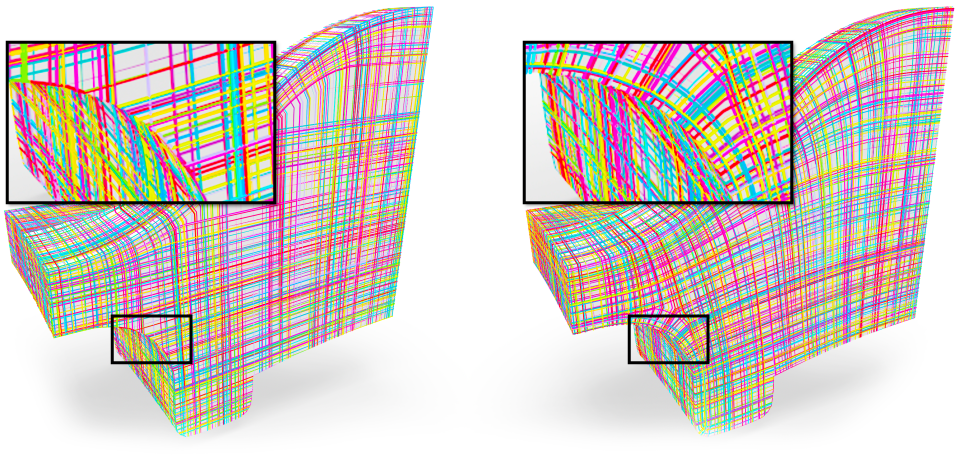}
        \put(0, -3){\textbf{(a) w/o feature line const.}}
        \put(54.5,-3){\textbf{(b) w/ feature line const.}}
	\end{overpic}
 \vspace{1mm}
 \caption{
 NeurCross supports feature line constraints. (a) The results without~(w/o) the feature line constraint~(const.);  and (b) The result with~(w/) the feature line constraint.
 }
 \label{fig:featureLine}
\end{figure}

\begin{figure}[t]
    \centering
    \graphicspath{{figures/}}
    \begin{overpic}
		[width=0.95\linewidth]{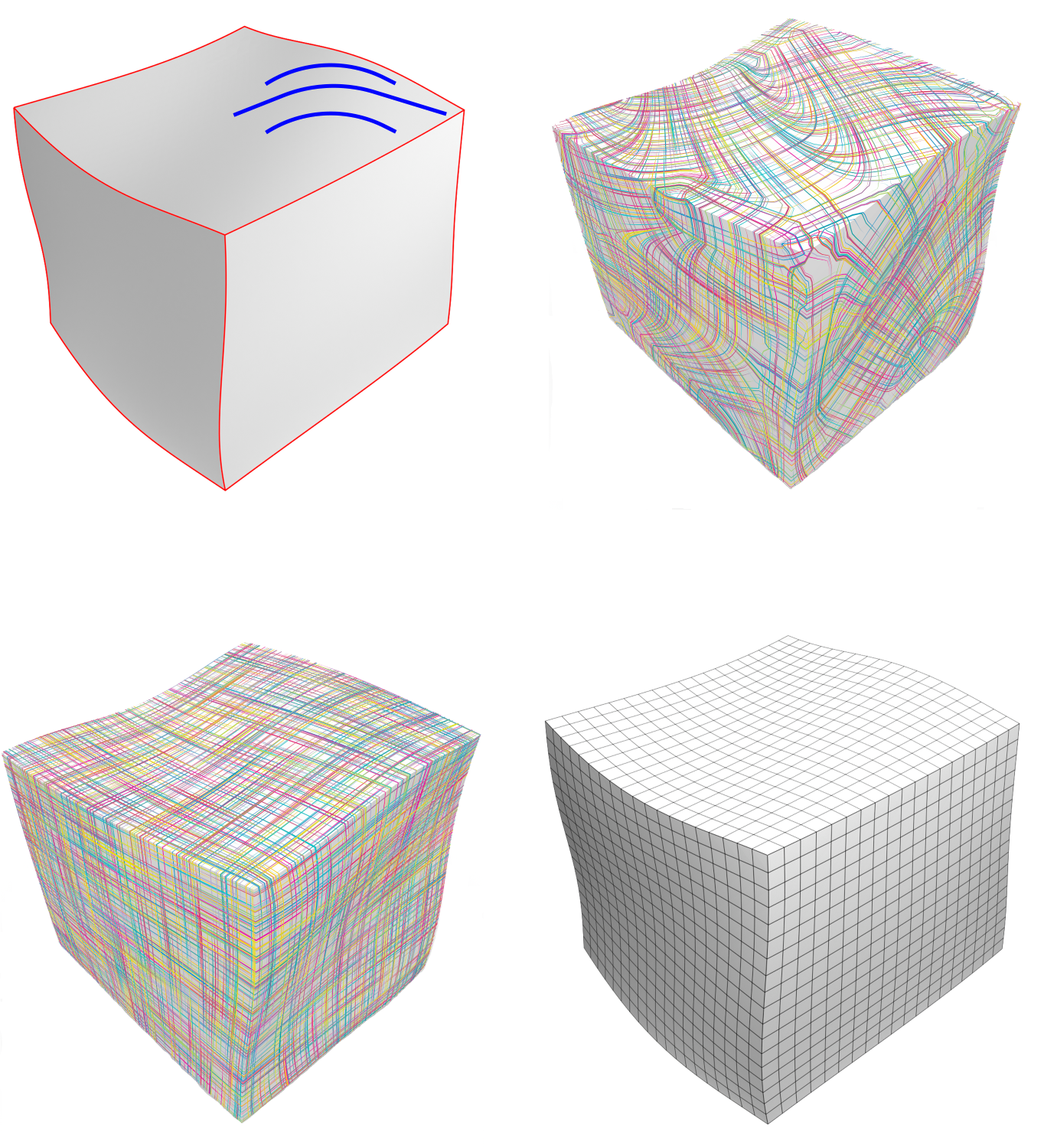}
        \put(8, 51.5){\textbf{(a) Input mesh}}
        \put(48,51.5){\textbf{(b) Principal curvature field}}
        \put(8, -2.8){\textbf{(c) Our cross field}}
        \put(53,-2.8){\textbf{(d) Our quad mesh}}
	\end{overpic}
 \vspace{1mm}
 \caption{
 NeurCross enforces alignment with sharp features even when they diverge from principal curvature directions.
(a) An input mesh with conflicting principal curvature directions (blue) and feature curves (red);  
(b) The principal curvature field of the input surface; 
(c) The cross field computed by NeurCross;
(d) The resulting quad mesh generated using NeurCross.}
 \label{fig:featureLine_curvature}
\end{figure}

As illustrated in the inset figure, let \( FL \) be the feature lines, where the cross field at each point of \( FL \) has been specified.
We use \( d_g(\boldsymbol{p}, FL) \) to denote the geodesic distance between a surface point \( \boldsymbol{p} \) and \( FL \). We introduce  
\begin{equation}  
\boldsymbol{D}_{\boldsymbol{p}} = 1 - \exp\left(-\rho_{\text{feature}} \, d_g(\boldsymbol{p}, FL)\right),  
\end{equation}  
and redefine the principal curvature direction alignment as follows:  
\begin{equation}  
\mathcal{L}_\text{AP} = \frac{1}{|\mathcal{P}|} \int_{\mathcal{P}} \boldsymbol{D}_{\boldsymbol{p}} \left( \big\vert \boldsymbol{H}_{\boldsymbol{p}} \boldsymbol{\alpha}_{\boldsymbol{p}} \times \boldsymbol{\alpha}_{\boldsymbol{p}} \big\vert + \big\vert \boldsymbol{H}_{\boldsymbol{p}} \boldsymbol{\beta}_{\boldsymbol{p}} \times \boldsymbol{\beta}_{\boldsymbol{p}} \big\vert \right) \text{d}\boldsymbol{p},  
\end{equation}
where \( \rho_{\text{feature}} \) (10 by default) in \( \boldsymbol{D}_{\boldsymbol{p}} \) serves as a sufficiently large constant to regulate the influence of the feature line. 
The color gradient in the insert figure represents the value of \( \boldsymbol{D}_{\boldsymbol{p}} \), which increases with distance from the feature line, reflecting the gradual reduction in \( FL \) influence on the surface point.
For ease of implementation, we approximate \( d_g(\boldsymbol{p}, FL) \) using straight-line distances. As shown in Fig.~\ref{fig:featureLine}, the crease line of the model is faithfully preserved in the neighborhood of \( FL \), while its influence diminishes as the distance increases.
Moreover, in regions where sharp features conflict with principal curvature directions, our NeurCross prioritizes feature alignment~(see Fig.~\ref{fig:featureLine_curvature}).

\paragraph{Rotation Angle Prediction}
Drawing inspiration from various object segmentation works~\cite{PDMeshNet2020, Hu2021SubdivNet}, we employ the U-Net architecture \cite{UNet2015} to construct our rotation angle prediction network. 
The input to our U-Net-based network includes the point cloud $\mathcal{P}$, along with the normal direction for each point, and the direction vectors $\boldsymbol{\nu}$ and $\boldsymbol{\mu}$ that represent the local coordinate system. The network outputs a scalar value $\omega_{\boldsymbol{p}} \in [0, 1]$ for each point $\boldsymbol{p}$, allowing the rotation angle $\theta_{\boldsymbol{p}}$ to be represented as $\theta_{\boldsymbol{p}} = 2\pi\omega_{\boldsymbol{p}}$.
For this module, we initialize the orientation at a point $\boldsymbol{p}$
using a normal distribution with a mean of 0 and a standard deviation of 0.2.

\begin{figure}[t]
    \centering
    \graphicspath{{figures/}}
    \begin{overpic}
		[width=0.95\linewidth]{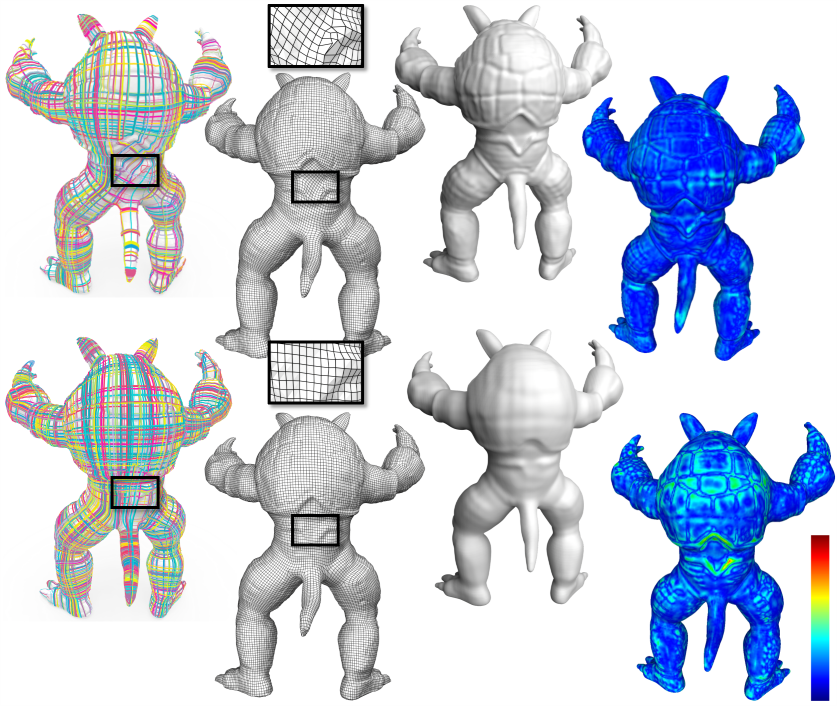}
            \put(0,53){\begin{turn}{90}{\textbf{Two-step}} \end{turn}}
        \put(0,20){\begin{turn}{90}{\textbf{Joint}}\end{turn}}
        \put(100,0){\begin{turn}{90}\small{\textbf{0}}\end{turn}}
        \put(100,14.5){\begin{turn}{90}\small{\textbf{0.01}}\end{turn}}
        \put(2.3,-5){\textbf{\small{(a) Cross field}}}
        \put(26,-5){\textbf{\small{(b) Quad mesh}}}
        \put(52,-5){\textbf{\small{(c) SDF fitting}}}
        \put(76,-5){\textbf{\small{(d) Fitting error}}}
	\end{overpic}
 \vspace{2.8mm}
 \caption{Comparison between the two-step method (top row) and our joint optimization strategy (bottom row).  
(a) Cross field; (b) Quad mesh; (c) The underlying SDF surface; (d) Fitting error between the SDF and the input triangular mesh. The final quad mesh is extracted based on the computed cross field and the input triangular mesh. As shown, our joint optimization strategy balances the overall smoothness of the cross field with alignment to the principal curvature directions.
}
\label{fig:step_by_step_optimization}
 \vspace{-4mm}
\end{figure}

\subsection{SDF and Cross Field Joint Optimization}
One challenge in quad meshing is balancing the overall simplicity of the cross field with alignment to the principal directions, a difficulty that becomes more pronounced for geometrically or topologically complex shapes. In this paper, we address this challenge by using an optimizable neural SDF as a bridge to achieve this balance. Notably, the neural SDF and the cross field are optimized simultaneously.

An alternative approach is to first fully optimize the SDF to accurately represent the input shape and then keep it fixed. However, in this case, the cross field may become severely constrained, as it must align with potentially irregular curvature lines of the pre-fixed SDF. 
As shown in Fig.~\ref{fig:step_by_step_optimization}, the fixed SDF, while providing an accurate representation, may introduce overly complex curvature lines, leading to an excessive number of singular points in the final cross field.

\subsection{Implementation Details}
\label{sec:imple_details}
\paragraph{SDF Loss Terms}
The loss terms used to regularize the SDF include the Eikonal condition~\cite{IGR}, the Dirichlet condition~\cite{phase}, and the alignment condition~\cite{HessianZX, RA2024alignHessian}. For further details on these loss terms, we refer readers to the existing literature~\cite{HessianZX, RA2024alignHessian, Dong2024NeurCADRecon}.

\paragraph{Sampling Strategy}
A neural SDF is employed to approximate the base surface, with regularization at sample points, as detailed in previous works \cite{HessianZX, RA2024alignHessian, Dong2024NeurCADRecon, IGR, SIREN, Rui2022RFEPS, SPSR, POCO, GalerkinNG, DiGS, NeuralPull, iPSR}. We extract centroids from all triangles in the mesh to define the sample set $\mathcal{P}$, chosen for their representative nature of the surface. For each point $\boldsymbol{p} \in \mathcal{P}$, the normal vector $\boldsymbol{n}_{\boldsymbol{p}}$ is derived from the corresponding triangular face's normal.

Given that each triangular face has three neighboring faces, neighboring relationships between points in $\mathcal{P}$ can be thus established. The SDF is assumed differentiable within a narrow, thin-shell space $\Omega$, which closely encloses the base surface. Following previous studies \cite{HessianZX, RA2024alignHessian, Dong2024NeurCADRecon, NeuralPull, IGR}, $\Omega$ is sampled using random displacements around each point $\boldsymbol{p} \in \mathcal{P}$. A Gaussian distribution centered at each $\boldsymbol{p}$, with a standard deviation based on the distance to its $k$-th nearest neighbor (typically $k=50$), is used for this purpose. A one-point sampling technique is then applied to generate the sample set $\Omega$, which is the same size as $\mathcal{P}$.

To prevent outlier zero iso-surfaces far from $\mathcal{P}$, we uniformly sample the bounding box (assuming input points are normalized within $[-0.5, 0.5]^3$), generating a sample set $\mathcal{Q}$.

\paragraph{Eikonal Condition}
The Eikonal condition is crucial for ensuring that the SDF \( f(\boldsymbol{x};\Theta) \) maintains a unit gradient at every point, i.e., \( \|\nabla f\| = 1 \), particularly in the vicinity of the surface. The corresponding loss term is defined as:
\begin{equation}
\mathcal{L}_{E} = \frac{1}{|\mathcal{P}| + |\Omega|} \int_{\mathcal{P} \cup \Omega} \big| 1 - \|\nabla f(\boldsymbol{x};\Theta)\| \big| \, \text{d}\boldsymbol{x},
\end{equation}
where \(\mathcal{P}\) represents the sample points (e.g., centroids of mesh triangles), and \(\Omega\) encodes a narrow band around the surface where the SDF is differentiable. Notably, the point set \(\mathcal{Q}\), which represents regions far from the base surface, is excluded from this loss term and serves to prevent outlier zero isosurfaces in distant regions.

\paragraph{Dirichlet Condition}
For every point \(\boldsymbol{p} \in \mathcal{P}\), it is essential that they lie as close as possible to the underlying surface, ideally satisfying \(f(\boldsymbol{p}; \Theta) = 0\). Conversely, for points \(\boldsymbol{q} \in \mathcal{Q}\), which lie away from the underlying surface, we aim to partition \(\mathcal{Q}\) into interior and exterior regions, preventing \(f\) from degenerating. These conditions are formalized as the following loss terms:
\begin{equation}
\mathcal{L}_\text{DM} = \frac{1}{|\mathcal{P}|} \int_{\mathcal{P}} \big| f(\boldsymbol{p}; \Theta) \big| \, \text{d}\boldsymbol{p},
\end{equation}
and
\begin{equation}
\mathcal{L}_\text{DNM} = \frac{1}{|\mathcal{Q}|} \int_{\mathcal{Q}} \exp\left(-\rho_{\text{DNM}} \big| f(\boldsymbol{q}; \Theta) \big|\right) \, \text{d}\boldsymbol{q},
\end{equation}
where \(\rho_{\text{DNM}}\) (defaulting to 100) is the exponential weight controlling the penalty for deviations from the surface.

\begin{figure}
    \centering
    \graphicspath{{figures/}}
    \begin{overpic}
		[width=0.98\linewidth]{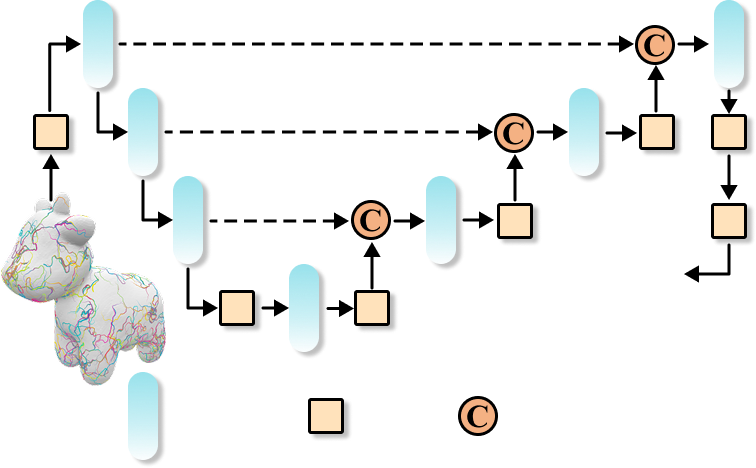}
  
		\put(30,46){Skip Connection}
		\put(83,22){{\fontsize{30}{0}\selectfont $\theta$}}
        \put(24,6){ResNets}
		\put(48,6){FC}
		\put(67.5,6){Concatenation}
  
	\end{overpic}
  \vspace{-1em}
 \caption{
 An overview of our U-Net-based module designed for predicting the rotation angle $\theta$. The network architecture incorporates the ResNet structure, with all layers being Multi-Layer Perceptrons (MLPs). The ``ResNets'' represents a combination of multiple ResNet blocks, the ``FC'' denotes the Fully Connected Layer, and the circled ``C'' symbol indicates the concatenation operation.
 }
 \label{fig:unet}
\end{figure}

\paragraph{SIREN-based Module}
Similar to various implicit surface reconstruction methods~\cite{IDF, phase, DiGS, HessianZX, RA2024alignHessian}, our NeurCross employs the SIREN~\cite{SIREN} network architecture, which consists of four hidden layers with 256 units each. The SIREN architecture is based on multi-layer perceptrons (MLPs), where inputs are first normalized to the range \([-1, 1]^3\) before being processed by the network. The activation function used in this architecture is the sine periodic function, which operates on the input point cloud \(\mathcal{P}\) to produce the SDF field required for computing the Hessian matrix. For initializing this SIREN-based module, we follow SIREN's initialization strategy~\cite{SIREN}, which ensures that the distribution of activations remains consistent across all layers of the network.

\paragraph{U-Net-based Module}
We adopt a U-Net architecture~\cite{UNet2015} as the backbone for predicting rotation angles. To address the vanishing gradient issue in deep networks, we incorporate ResNet blocks~\cite{He2016ResNet} as the core components of the U-Net. As shown in Fig.~\ref{fig:unet}, our network consists of ResNet blocks (ResNets) and fully connected layers (FC), with all layers implemented using MLPs. A detailed configuration of each ResNet block is provided in Tab.~\ref{tab:unet}.

\newcommand{\blockb}[3]{\multirow{3}{*}{\(\left[\begin{array}{c}\text{1$\times$1, #1}\\[-.1em] \text{1$\times$1, #2}\\[-.1em] \text{1$\times$1, #2}\end{array}\right]\)$\times$#3}
}
\renewcommand\arraystretch{1.1}
\begin{table}[tb]
  \caption{
The sizes of the building blocks within our U-Net-based module.
$n_{\text{bottleneck}}$ denotes the number of bottleneck layers within each ResNet block. From the 1st to the 7th block, the values of $n_{\text{bottleneck}}$ are set to 3, 4, 6, 3, 3, 4, and 6, respectively.
}
  \label{tab:unet}
\begin{center}
\resizebox{0.9\linewidth}{!}{
  \begin{tabular}{c|c|c}
  \toprule
  \textbf{Layer Name} & \textbf{Layer Architecture} & \textbf{Output Size}\\
  \hline
    & 1$\times$1, input\_size=12 &  256 \\
  \hline
  \multirow{3}{*}{ResNets \#1, \#2, \#3} &  \blockb{256}{64}{$n_\text{bottleneck}$} & \multirow{3}{*}{256}\\
  &  &\\
  &  &\\
  \hline
  \multirow{3}{*}{ResNets \#4, \#5, \#6, \#7} & \blockb{512}{128}{$n_\text{bottleneck}$} & \multirow{3}{*}{512} \\
  &  &\\
  &  &\\
  \hline
  & 1$\times$1, 512 &  32 \\
  \hline
   & 1$\times$1, 32  & 1 \\
  \bottomrule
  \end{tabular}}%
  \end{center}
\end{table}

\paragraph{Parameter Setting}
In this paper, we set the weights as follows based on our tailored configurations: $\lambda_\text{E} = 50$, $\lambda_{\text{DM}} = 7000$, $\lambda_{\text{DNM}} = 600$, $\lambda_{\text{AN}} = 3$, $\lambda_{\text{AP}} = 10$, and $\lambda_{\text{S}} = 30$. The annealing factor $\tau$ remains 1 during the initial 20\% of iterations, then linearly decreases to $3 \times 10^{-4}$ from 20\% to 40\% of the iteration span, and finally drops to 0 towards the end. Throughout the training phase, we apply the Adam optimizer~\cite{Adam} with a default learning rate of $5 \times 10^{-5}$ and complete 10,000 iterations.

\paragraph{Quad Mesh Extraction}
The extraction of the quad mesh from our cross field follows a two-step scheme, as detailed in references \citet{MIQ2009}, \citet{libQEX13}, and~\citet{DL2quadMesh2021}, to achieve a high-quality outcome. This process begins with a step of parametrization based on our cross field. In implementation, we utilize the global-seamless parametrization technique from libigl~\cite{libigl2017} to align the parametrization with our cross field. Subsequently, we employ libQEx \cite{libQEX13} to extract the quad mesh from this parameterization.

\begin{figure*}[!t]
    \centering
    \graphicspath{{figures/}}
    \begin{overpic}
		[width=\linewidth]{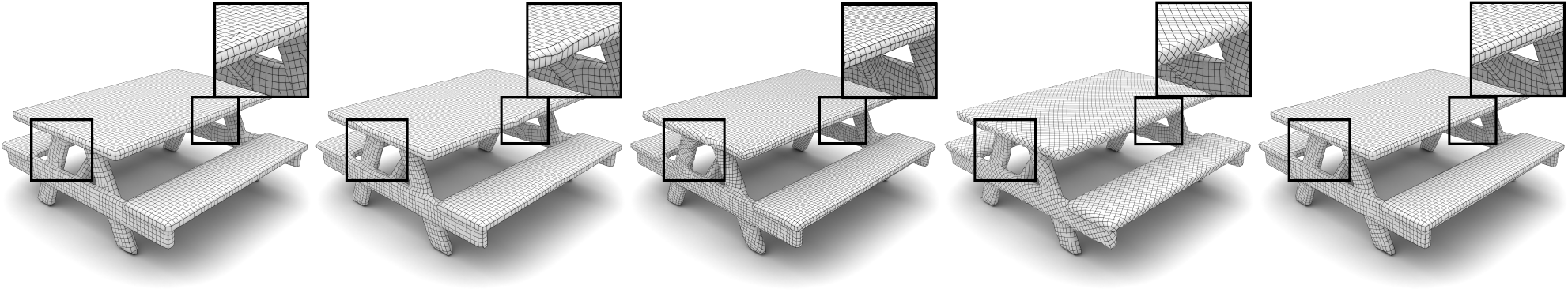}

        \put(9.1,-1.7){\textbf{IM}}
        \put(26.7,-1.7){\textbf{QuadriFlow}}
        \put(47.0,-1.7){\textbf{QuadWild}}
        \put(68.8,-1.7){\textbf{MIQ}}
        \put(84,-1.7){\textbf{NeurCross (Ours)}}
	\end{overpic}
   \vspace{-3mm}
 \caption{Quad meshes generated by NeurCross and four other methods on the table model in the ShapeNet dataset~\cite{ShapeNet}.}
 \label{fig:ShapeNet}
 \vspace{-2mm}
\end{figure*}

\section{Experiments}
\paragraph{Evaluation Metrics and Platform}
To evaluate the accuracy of the quad mesh, we utilize four primary metrics~\cite{QuadriFlow2018, HessianZX}: area distortion~(Area), angle distortion~(Angle), the number of singularities~(\# of Sings), chamfer distance~(CD), and Jacobian Ratio~(JR). 
The area distortion metric, scaled by 10,000, represents the standard deviation of the areas of the quadrilateral faces within a mesh.
The angle distortion is quantified using the formula ~$\sqrt{\frac{1}{N}\sum_i(\phi_i - \frac{\pi}{2})^2}$, where the summation extends over all angles~$\phi$ in the quad mesh, and~$N$ denotes their count.
Chamfer distance, scaled by 10,000 and calculated using the $L_1$-norm, quantifies the similarity between two surfaces.
The Jacobian Ratio quantifies the uniformity of local deformation in quadrilateral elements. It is defined as the ratio of the smallest to the largest determinant of the Jacobian matrices at element corners, providing a dimensionless measure from 0 (degenerate element) to 1 (perfect parallelogram).
The experiments detailed in this paper were executed on an NVIDIA GeForce RTX 3090 graphics card equipped with 24GB of video memory and powered by an AMD EPYC 7642 processor.

\paragraph{Datasets}
We carry out quad mesh generation experiments on two popular datasets: ShapeNet~\cite{ShapeNet} and Thingi10K~\cite{Thingi10K}.
To maintain uniformity in evaluation, all input meshes are scaled to fit within the range of~$[-0.5, 0.5]^3$ ensuring a consistent and fair basis for comparison across all datasets.

\subsection{Comparison on Open Datasets}
\label{sec:Comparison}
We assess the efficacy of our proposed method, NeurCross, by conducting evaluations on two distinct datasets and comparing its performance against four contemporary state-of-the-art quadrilateral mesh generation methods.
For the three methods, namely Instant Meshes (IM) \cite{Instant_Meshes2015}, QuadriFlow \cite{QuadriFlow2018}, and QuadWild~\cite{quadwild2021}, we employed the open-source implementations that are readily available. 
It is worth noting that QuadWild~\cite{quadwild2021} is primarily a quadrangulation method rather than a cross field generation approach.
The Mixed-Integer Quadrangulation (MIQ) method~\cite{MIQ2009} does not release its source code; therefore, we use the implementation provided by libigl~\cite{libigl2017}. However, the available implementation does not support the feature alignment constraint. For a fair comparison with MIQ, we employ the same parameterization and extraction techniques, namely global-seamless parameterization and libQEx~\cite{libQEX13}.

\begin{table}[!t]
\vspace{1.5mm}
\centering
\caption{Quantitative comparison on the ShapeNet dataset~\cite{ShapeNet}. Within each column, the best scores are emphasized with bold and underlining (\underline{\textbf{best}}), whereas the second-best scores are highlighted in bold (\textbf{second best}). 
The quad mesh generated by all the methods comprises an average of 6,000 vertices and 12,000 faces.}
\label{tab:shapenet}
\resizebox{0.98\linewidth}{!}{
\begin{tabular}{l|ccccc} 
\toprule
 & Area~$\downarrow$ & Angle~$\downarrow$ & \# of Sings~$\downarrow$ & CD~$\downarrow$ & JR~$\uparrow$\\
\midrule
IM~\cite{Instant_Meshes2015}  & 1.57 & 11.78 & 200.52 & 8.97 & 0.70\\
QuadriFlow~\cite{QuadriFlow2018} & 2.28 & 13.24 & 91.58 & 50.18 & 0.65\\
QuadWild~\cite{quadwild2021} & \textbf{1.52} & \textbf{11.05} & 93.04 & 10.34 & \textbf{0.73}\\
MIQ~\cite{MIQ2009} & 5.23 & 12.89 & \underline{\textbf{82.12}} & \textbf{8.25} & 0.58\\
\midrule
\textbf{NeurCross (Ours)}  & \underline{\textbf{1.48}} & \underline{\textbf{9.85}} & \textbf{85.32} & \underline{\textbf{8.03}} & \underline{\textbf{0.78}}\\
\bottomrule
\end{tabular}
}
\vspace{-2mm}
\end{table}

\begin{figure*}
    \centering
    \graphicspath{{figures/}}
    \begin{overpic}
		[width=\linewidth]{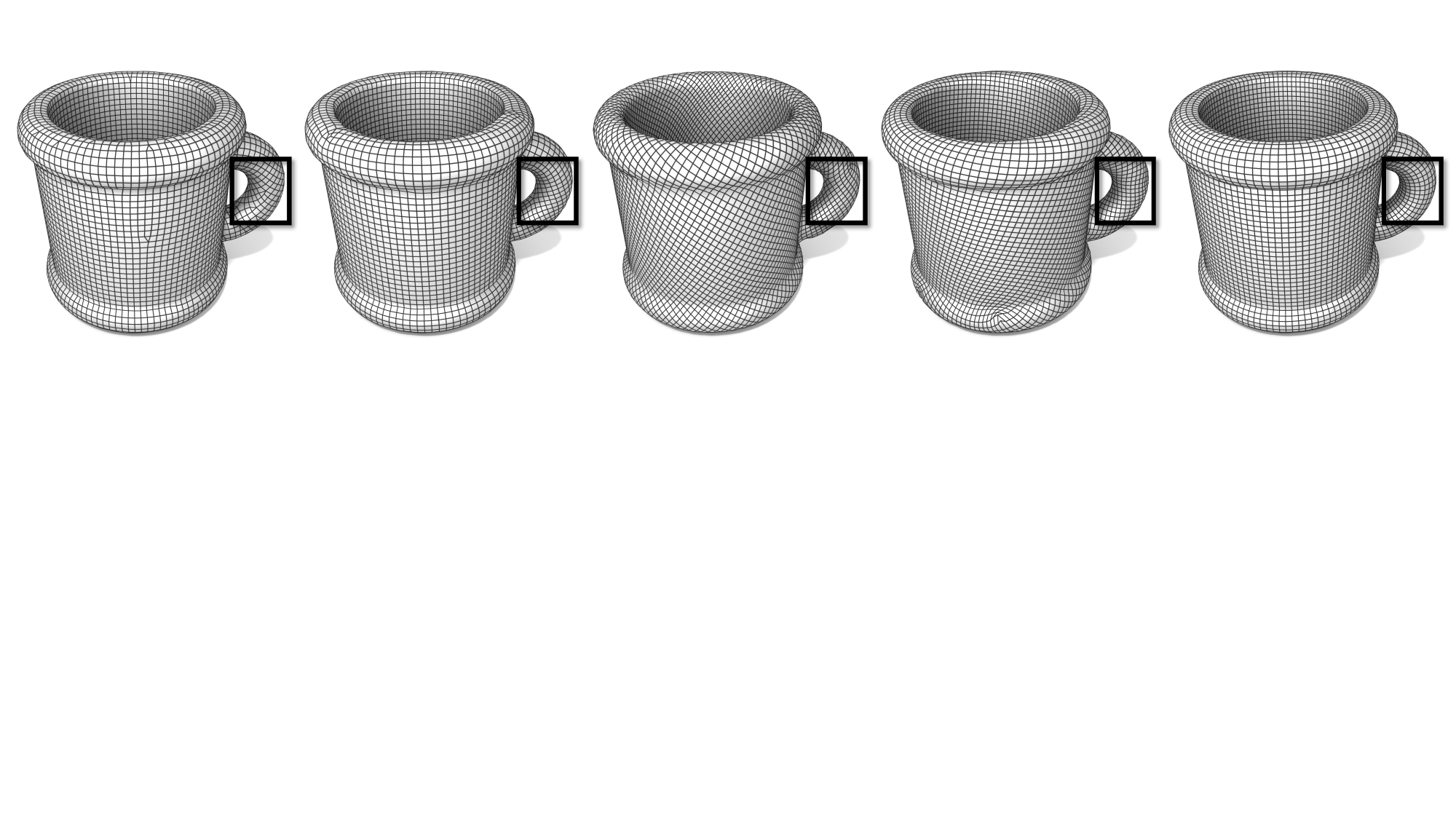}

        \put(7.5,-2.3){\textbf{IM}}
        \put(25,-2.3){\textbf{QuadriFlow}}
        \put(45,-2.3){\textbf{QuadWild}}
        \put(67.5,-2.3){\textbf{MIQ}}
        \put(83,-2.3){\textbf{NeurCross (Ours)}}
	\end{overpic}
   \vspace{-1.5mm}
\caption{Quad meshes generated by NeurCross and four other methods on a cup model in the Thingi10K dataset~\cite{Thingi10K}.}
 \label{fig:Thingi10k}
\end{figure*}

\paragraph{ShapeNet Dataset}
The ShapeNet dataset~\cite{ShapeNet} consists of a diverse range of human-made models. As the global-seamless parametrization from libigl~\cite{libigl2017} cannot handle non-manifold meshes, we use manifold ShapeNet meshes repaired with DualOctreeGNN~\cite{DualOctreeGNN2022}. We apply our NeurCross to three randomly selected categories—airplane, bench, and cabinet—which together contain 7433 models. For a fair comparison, all generated quad meshes are standardized to contain an average of 6,000 vertices and 12,000 faces.

In Fig.~\ref{fig:ShapeNet}, we display the quad meshes generated by our NeurCross alongside four other methods. 
For this example, although IM~\cite{Instant_Meshes2015} produces a regular quadrilateral mesh, the outcome includes some triangular elements. 
More comparisons between IM and our method will be provided in Sec.~\ref{sec:sings}. QuadriFlow~\cite{QuadriFlow2018}, MIQ~\cite{MIQ2009}, and QuadWild~\cite{quadwild2021} generate some misaligned quadrilateral elements, as seen in the highlighted windows. In contrast, our method yields a better quadrilateral mesh.
Tab.~\ref{tab:shapenet} shows the quantitative comparison of our method against the four approaches.

\begin{table}[!t]
\vspace{1.5mm}
\centering
\caption{Quantitative comparison on the Thingi10K dataset~\cite{Thingi10K}.
The quad mesh generated by all methods comprises an average of 10,000 vertices and 20,000 faces.
Within each column, the best scores are emphasized with bold and underlining (\underline{\textbf{best}}), whereas the second-best scores are simply highlighted in bold (\textbf{second best}).}
\label{tab:thingi10k}
\resizebox{0.98\linewidth}{!}{
\begin{tabular}{l|ccccc} 
\toprule
 & Area~$\downarrow$ & Angle~$\downarrow$ & \# of Sings~$\downarrow$ & CD~$\downarrow$ & JR~$\uparrow$\\
\midrule
IM~\cite{Instant_Meshes2015}  & 1.45 & 10.57 & 397.18 &  9.83 &  0.75\\
QuadriFlow~\cite{QuadriFlow2018} & 1.58 & 12.39 & 78.32 & 26.89 & 0.72\\
QuadWild~\cite{quadwild2021} & 1.40 & 10.16 & 85.11 & 28.12 & \textbf{0.77}\\
MIQ~\cite{MIQ2009} & \textbf{1.38} & \textbf{9.85} & \underline{\textbf{66.54}} & \textbf{8.57} & 0.67\\
\midrule
\textbf{NeurCross (Ours)}  & \underline{\textbf{1.33}} & \underline{\textbf{9.68}} & \textbf{68.96} & \underline{\textbf{8.22}} & \underline{\textbf{0.81}}\\
\bottomrule
\end{tabular}
}
\vspace{-2mm}
\end{table}

\begin{figure}[h]
    \centering
    \graphicspath{{figures/}}
    \begin{overpic}
		[width=7.5cm]{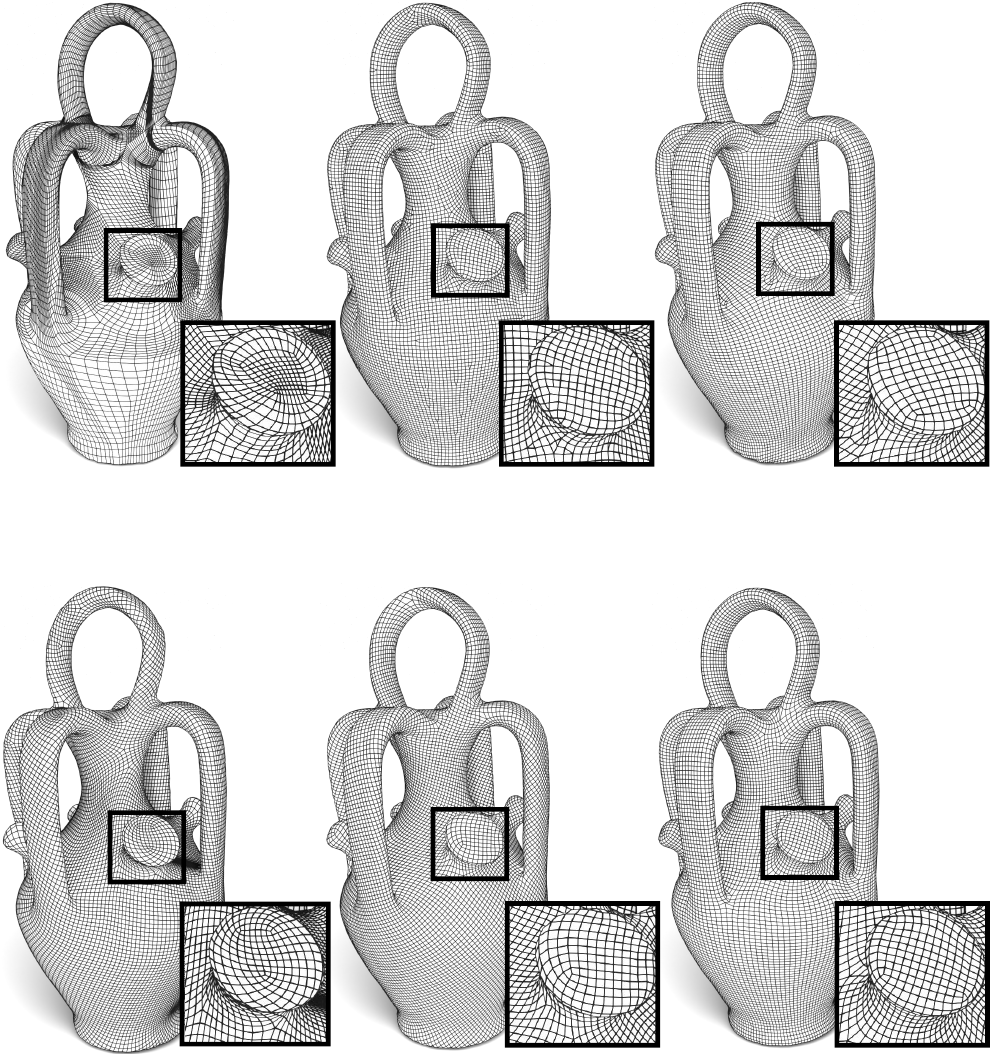}
		\put(8.5,49.8){\textbf{IGM}}
		\put(43,49.8){\textbf{IM}}
		\put(69,49.8){\textbf{QuadriFlow}}
        \put(6.7,-5.5){\textbf{QuadWild}}
		\put(42,-5.5){\textbf{MIQ}}
		\put(63,-5.5){\textbf{NeurCross (Ours)}}
	\end{overpic}
   \vspace{3mm}
 \caption{Comparison with five state-of-the-art methods using data provided in IGM~\cite{IGM2013}.}
 \label{fig:compare_IGM}
 \vspace{-2mm}
\end{figure}

\paragraph{Thingi10K Dataset}
The Thingi10K dataset~\cite{Thingi10K} features a variety of shapes with intricate geometric details. 
For our analysis based on Thingi10K, we tested 1,000 randomly selected triangle meshes from the dataset, which were also used as inputs for all comparative methods. The quad meshes generated by all methods contain, on average, 10,000 vertices and 20,000 faces to maintain fidelity to the original models.

Quantitative comparison statistics are presented in Tab.~\ref{tab:thingi10k}. Our method consistently outperforms others on average across this dataset. 
Interestingly, MIQ~\cite{MIQ2009} shows commendable performance on this dataset. However, it is important to note that despite this improvement, the issue of producing distorted quadrilaterals in the resulting quad mesh remains (see Fig.~\ref{fig:Thingi10k} and the JR metric in Tab.~\ref{tab:thingi10k}). 
Quadwild~\cite{quadwild2021} requires smoothing of the generated quad mesh, which compromises geometric details and increases the Chamfer Distance (CD). 
In contrast, our method produces a more intuitive cross field without needing to introduce excessive singular points (see zoom-in windows in Fig.~\ref{fig:Thingi10k}).

\begin{figure}
    \centering
    \graphicspath{{figures/}}
    \begin{overpic}
		[width=8cm]{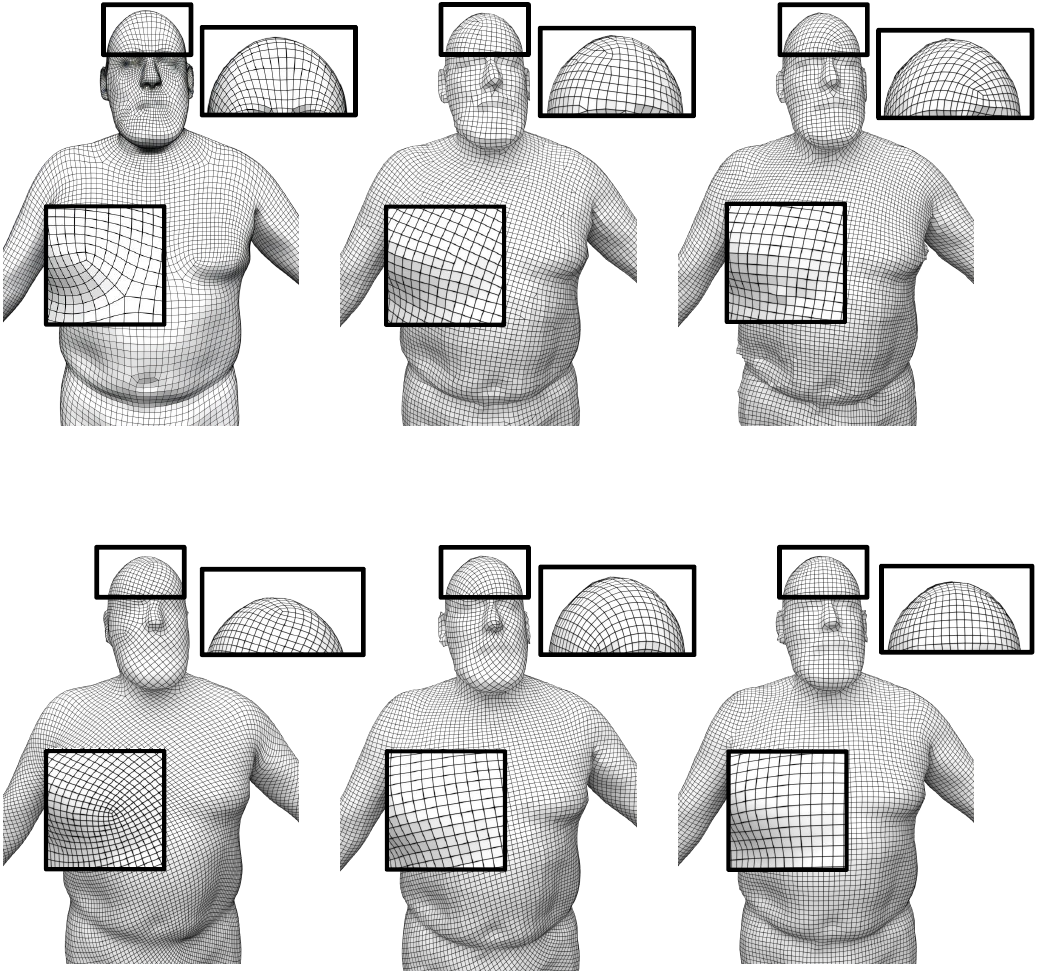}
	
		\put(5,47.5){\textbf{Dielen et al.}}
		\put(45.7,47.5){\textbf{IM}}
		\put(71.0,47.5){\textbf{QuadriFlow}}
        \put(5.44,-4.5){\textbf{QuadWild}}
		\put(44.5,-4.5){\textbf{MIQ}}
		\put(65,-4.5){\textbf{NeurCross (Ours)}}
	\end{overpic}
   \vspace{2mm}
 \caption{Comparison with five methods on Human Body data. \citet{DL2quadMesh2021} proposed a supervised learning-based approach designed to generate quad meshes on human body data from the FAUST dataset~\cite{FAUST}. Due to the absence of available open-source data, the comparison result in the upper left is taken from~\citet{DL2quadMesh2021}'s paper.
}
 \label{fig:compare_learning}
 \vspace{-2mm}
\end{figure}

\subsection{Further Comparison}
\label{sec:more_comparison}
\paragraph{Comparison with IGM}
IGM~\cite{IGM2013} is characterized as a global approach, primarily focused on the joint optimization of parametrization with integer constraints. Like our method, it also utilizes libQEx~\cite{libQEX13} for extracting quad meshes from the parametrization. Although IGM provides full control over edge alignment and singularity placement, yielding high-quality quad meshes, its lack of scalability can lead to severely distorted quadrilaterals (see Fig.~\ref{fig:compare_IGM}).

\paragraph{Comparison with Learning Methods}
\citet{DL2quadMesh2021} represents a pioneering effort in quad mesh generation through deep learning methodologies. Their method uses a supervised network architecture to predict the frame field, comprising both a global network and a local network for field prediction. Subsequently, the parametrization-based quadrangulation method proposed in \citet{RN610} is employed to generate the quad meshes. However, due to the inherent constraints of supervised learning, this approach shows optimal performance only on the FAUST dataset \cite{FAUST}, a limitation not encountered by our self-supervised method. Owing to a lack of required data, our comparison is limited to the model presented in their paper (see Fig.~\ref{fig:compare_learning}).

\begin{figure}[t]
    \centering
    \graphicspath{{figures/}}
    \begin{overpic}
		[width=0.95\linewidth]{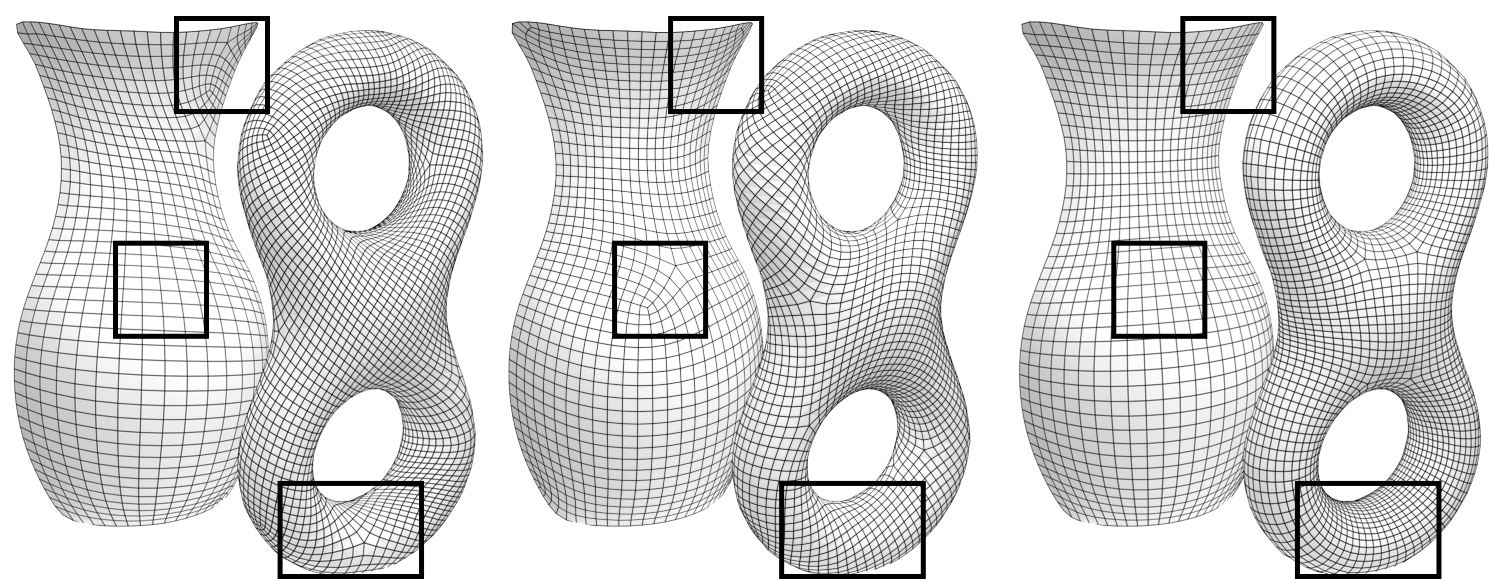}
		\put(6,-6){\textbf{Power Fields}}
        \put(40,-6){\textbf{PolyVectors}}
		\put(70,-6){\textbf{NeurCross (Ours)}}
	\end{overpic}
   \vspace{3.5mm}
   \caption{
    Comparison of quad meshes generated by Power Fields~\cite{PowerFields_Felix2013}, PolyVectors~\cite{PolyVectors_Diamanti2014}, and NeurCross.
}
 \label{fig:compare_powerFields_polyVectors}
 \vspace{-3mm}
\end{figure}

\begin{figure}
    \centering
    \graphicspath{{figures/}}
    \begin{overpic}
		[width=0.95\linewidth]{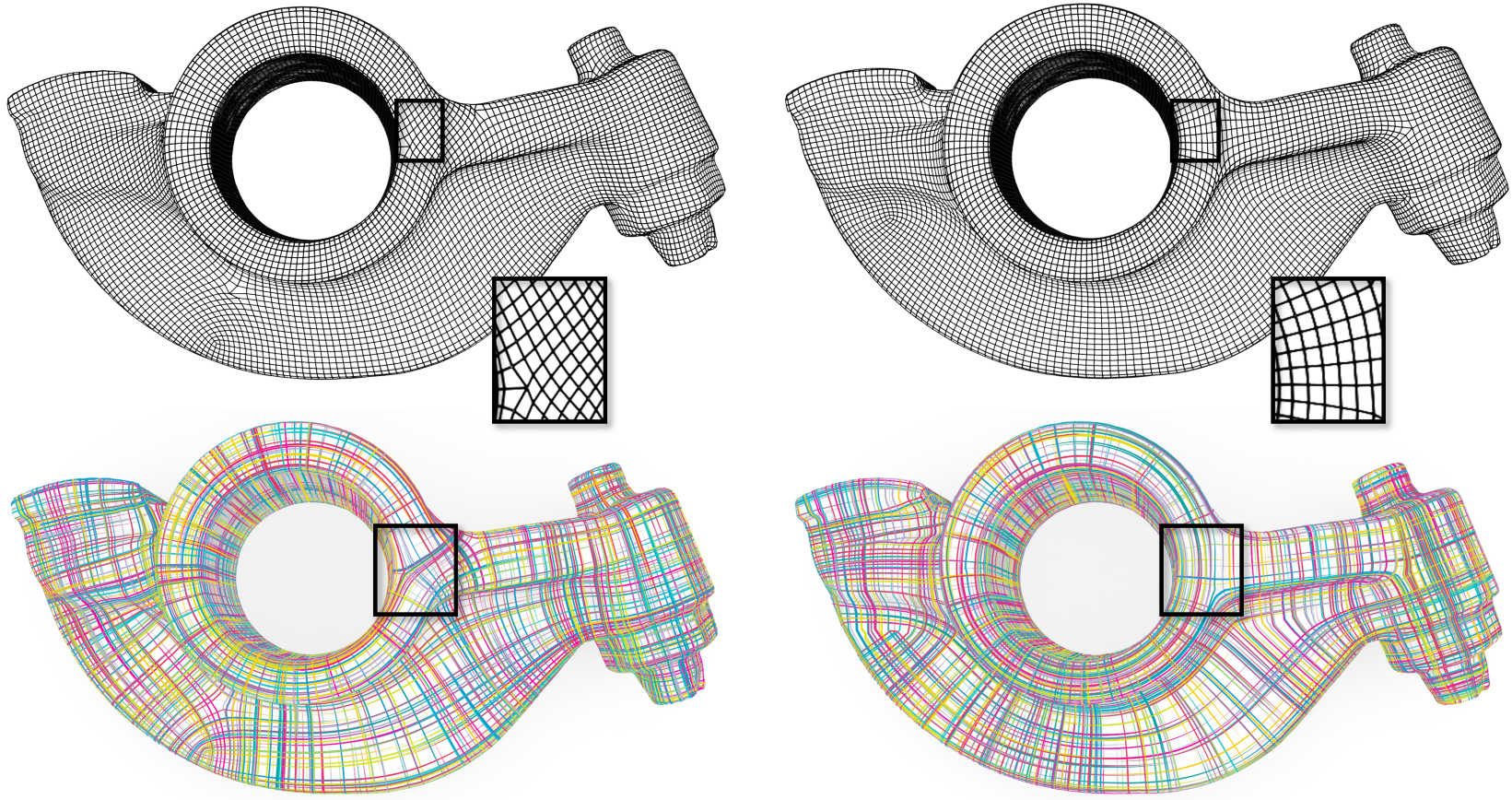}
        \put(20,-6){\textbf{IM}}
		\put(58,-6){\textbf{NeurCross (Ours)}}
	\end{overpic}
   \vspace{1em}
   \caption{
Comparison with IM. Here, the same approach—applying global seamless parameterization~\cite{libigl2017} and libQEx~\cite{libQEX13}— is used to extract quadrilateral meshes from the respective cross fields of IM and our NeurCross.
}
 \label{fig:compare_IM}
 \vspace{-3mm}
\end{figure}

\begin{figure}
    \centering
    \graphicspath{{figures/}}
    \begin{overpic}
		[width=0.95\linewidth]{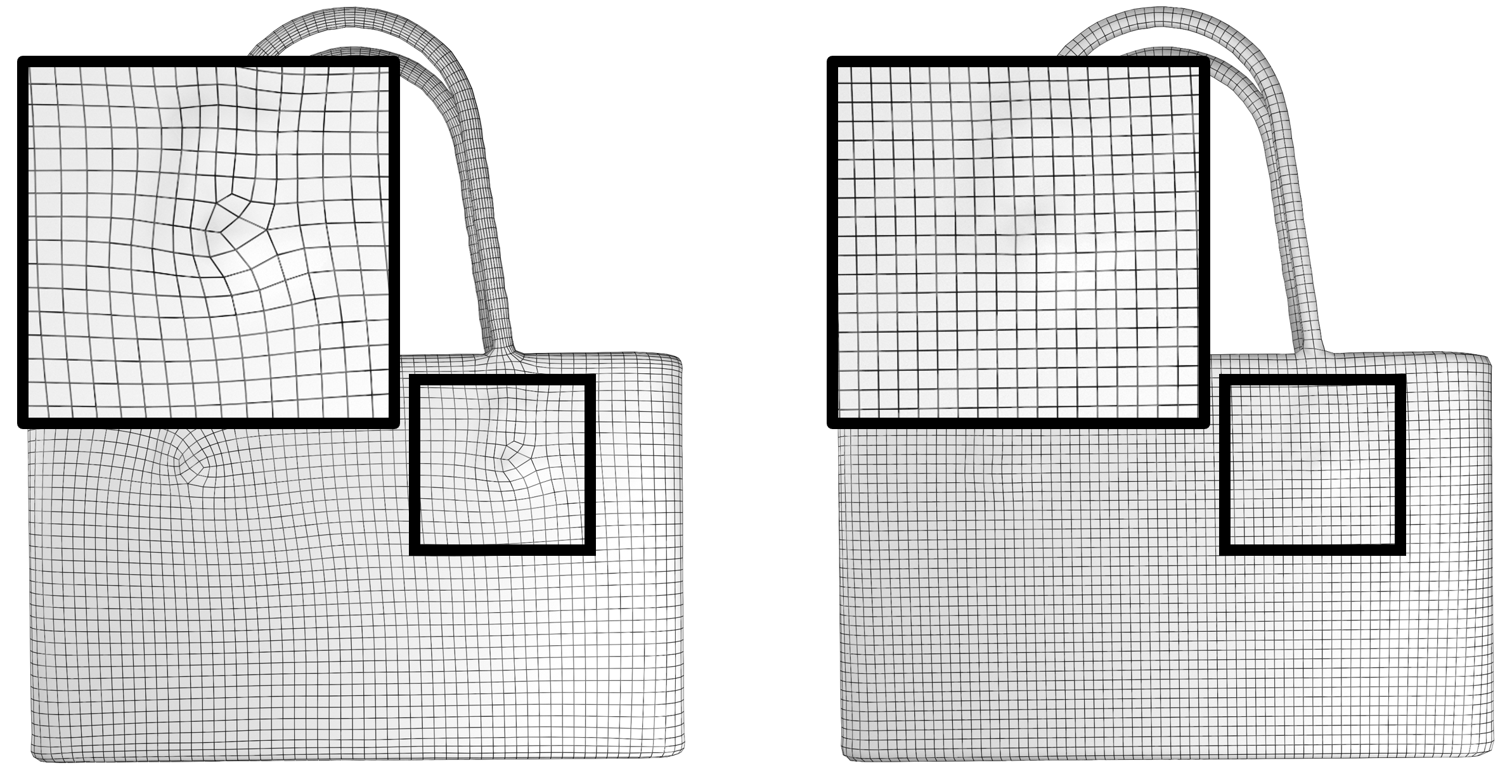}
		\put(12,-5){\textbf{Quad Remesher}}
		\put(62,-5){\textbf{NeurCross (Ours)}}
	\end{overpic}
   \vspace{3mm}
   \caption{
    Comparison of quad meshes generated by Quad Remesher~\cite{Quad_Remesher} and NeurCross.
}
 \label{fig:compare_quadmesher}
 \vspace{-3mm}
\end{figure}

\paragraph{Comparison with Power Fields and PolyVectors}
Power Fields~\cite{PowerFields_Felix2013} efficiently constructs smooth n-direction fields on surfaces by solving a sparse eigenvalue problem, ensuring global optimality and high-quality results. PolyVectors~\cite{PolyVectors_Diamanti2014} extends N-RoSy fields to N-PolyVector fields by relaxing orthogonality and symmetry constraints, enabling their computation via a sparse linear system without integer variables. Both methods focus on efficient computation of directional fields, with Power Fields~\cite{PowerFields_Felix2013} optimizing smoothness and PolyVectors~\cite{PolyVectors_Diamanti2014} generalizing traditional field representations.
Fig.~\ref{fig:compare_powerFields_polyVectors}
compares the quadrilateral meshes generated by our NeurCross and these methods. NeurCross not only aligns with principal curvatures but also preserves overall smoothness.

\paragraph{Comparison with IM}
IM~\cite{Instant_Meshes2015} is an effective method for generating quad meshes. To facilitate a fair comparison between IM and our approach, we use the same global seamless parameterization and extraction technique (libQEx~\cite{libQEX13}) to extract the quad mesh. As shown in Fig.~\ref{fig:compare_IM}, our method produces fewer singularities than IM. Additionally, our method outperforms IM~\cite{Instant_Meshes2015} in terms of principal direction alignment and structural integrity, as illustrated in the close-up views.

\paragraph{Comparison with Quad Remesher}
Quad Remesher~\cite{Quad_Remesher} excels at generating quadrilateral meshes and is available as a plugin for software like Blender. It is stable, efficient, and effective at preserving model features while maintaining topological uniformity, with our method achieving comparable results. However, its performance depends heavily on the quality of the input mesh, producing low-quality quadrilateral meshes when the input polygonal mesh is suboptimal (see Fig.~\ref{fig:compare_quadmesher}).

\begin{figure}[t]
    \centering
    \graphicspath{{figures/}}
    \begin{overpic}
		[width=7.5cm]{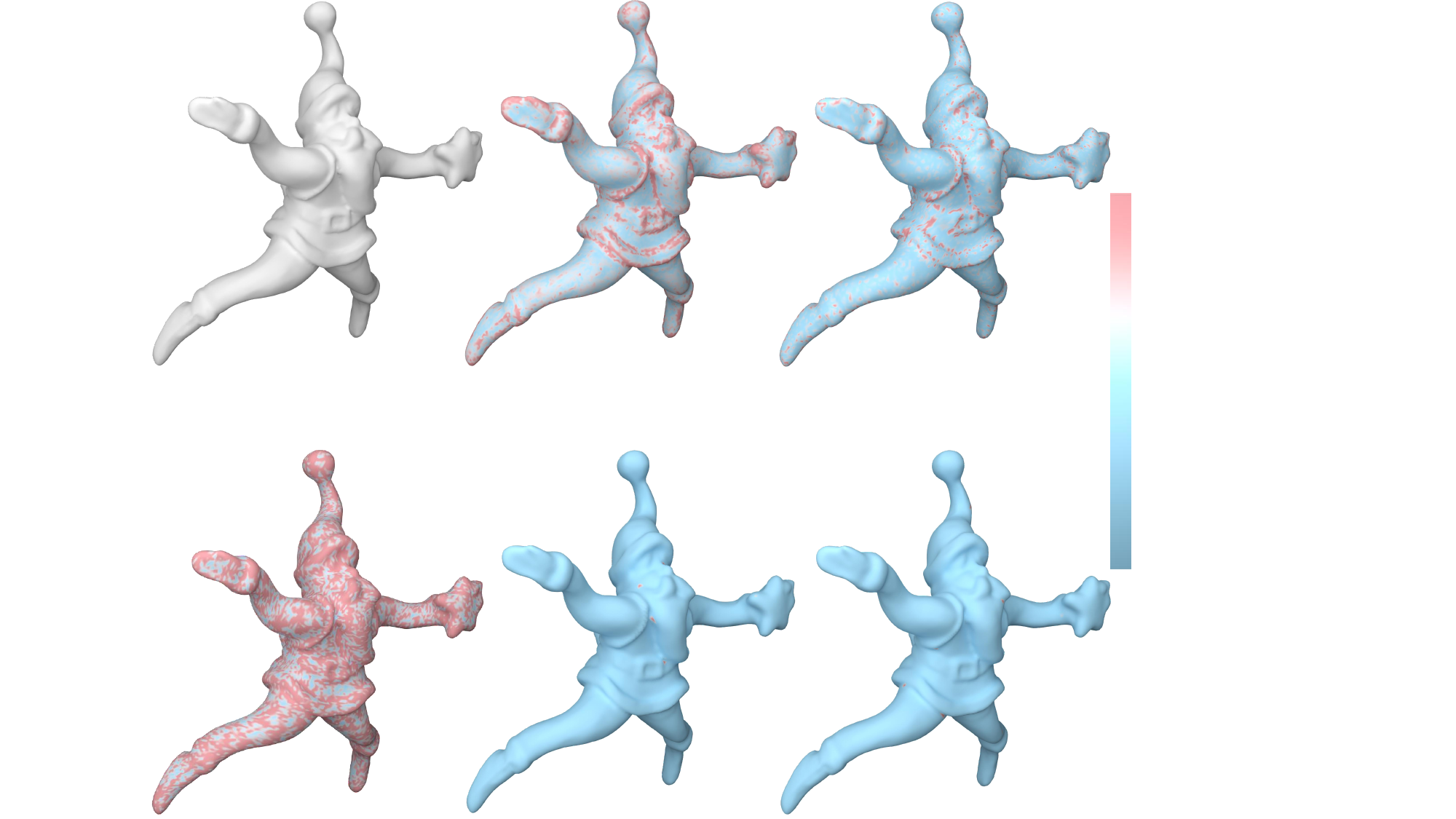}
	
		\put(8,42.5){\textbf{Input}}
		\put(45.5,42.5){\textbf{IM}}
		\put(67.8,42.5){\textbf{QuadriFlow}}
        \put(4,-4){\textbf{QuadWild}}
		\put(43,-4){\textbf{MIQ}}
		\put(65,-4){\textbf{NeurCross (Ours)}}
  
        \put(100,46.0){\begin{turn}{90}{$\boldsymbol{1.2\times10^{-4}}$} \end{turn}}
        \put(100,22){\begin{turn}{90}{\textbf{0}} \end{turn}}
	\end{overpic}
  \vspace{3mm}
 \caption{
 {\bf Approximation accuracy}. Here we show the approximation errors between the input surface and the final quad meshes generated by different methods. The error is measured from each sampled point on the quad mesh to the input surface.
 }
 \label{fig:hausdorff}
 \vspace{-5mm}
\end{figure}

\begin{figure*}[t]
    \centering
    \graphicspath{{figures/}}
    \begin{overpic}
		[width=0.95\linewidth]{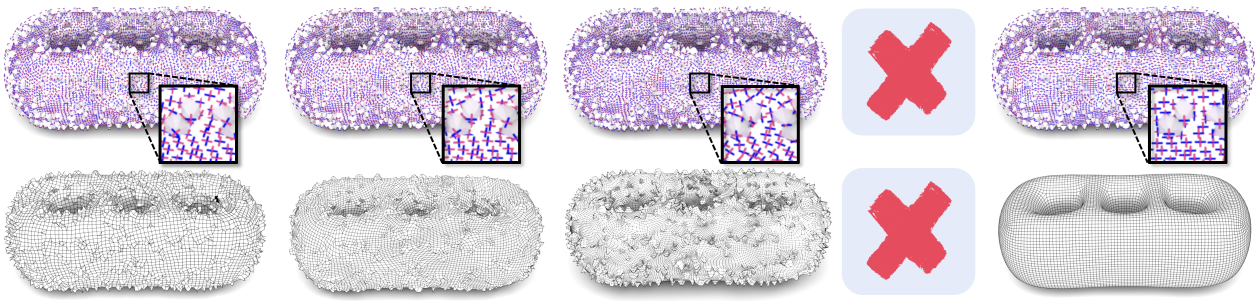}
        \put(10,-2.5){\textbf{IM}}
        \put(29,-2.5){\textbf{QuadriFlow}}
        \put(51,-2.5){\textbf{QuadWild}}
        \put(71,-2.5){\textbf{MIQ}}
        \put(82,-2.5){\textbf{NeurCross (Ours)}}
	\end{overpic}
  \vspace{3mm}
 \caption{
 The top row shows the cross field generated by our method and four other methods on a noisy input mesh. The bottom row shows the resulting quad meshes produced by each approach. Note that MIQ fails to produce a valid result for this input surface with noise.
 }
 \label{fig:noise}
 \vspace{2mm}
\end{figure*}

\begin{figure*}[t]
    \centering
    \graphicspath{{figures/}}
    \begin{overpic}
		[width=0.98\linewidth]{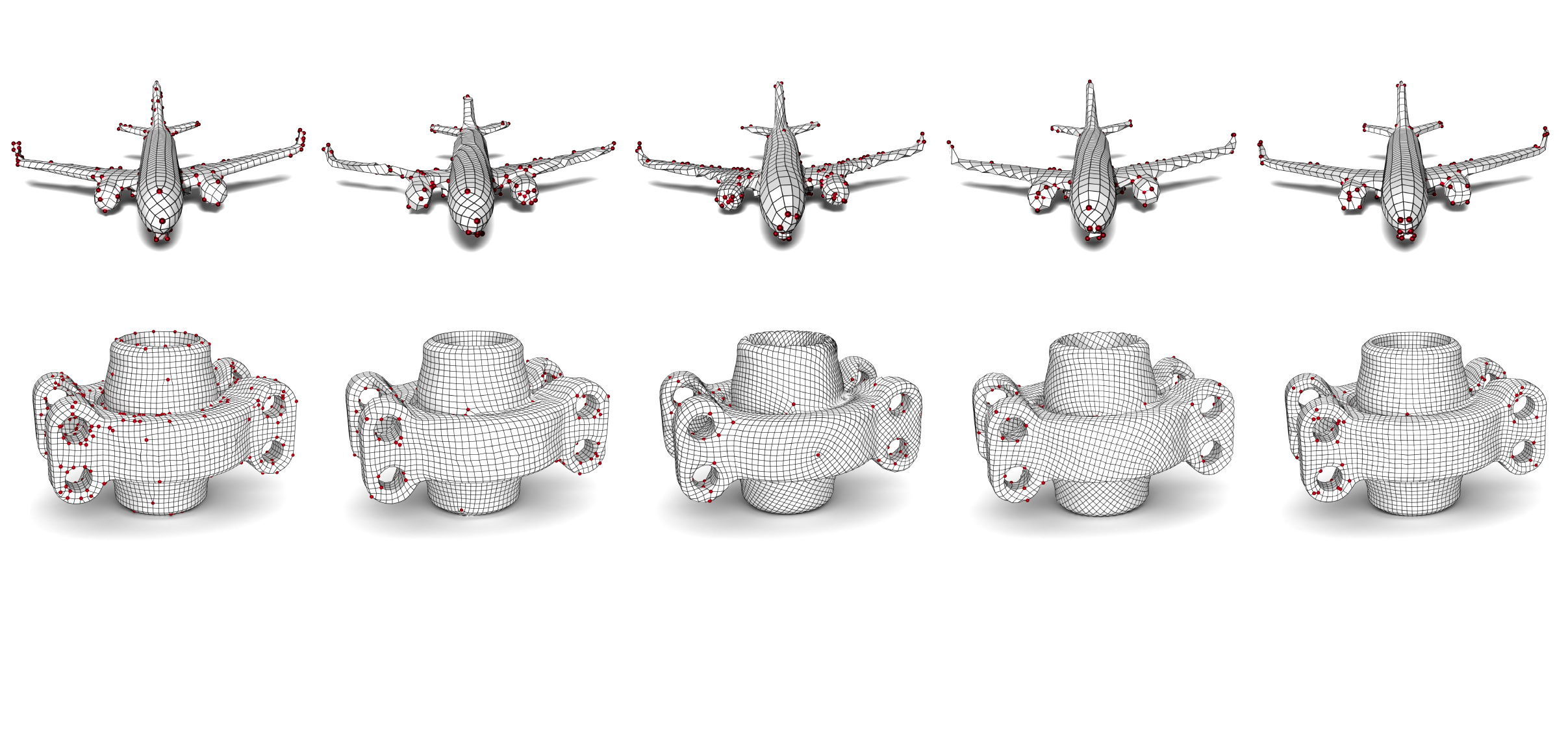}

        \put(8,31){\textbf{IM}}
        \put(25.5,30.5){\textbf{QuadriFlow}}
        \put(46.1,30.5){\textbf{QuadWild}}
        \put(68.8,30.5){\textbf{MIQ}}
        \put(83,30.5){\textbf{NeurCross (Ours)}}
        
		\put(5,17){\textbf{171~\# of Sings}}
        \put(5,-2.0){\textbf{443~\# of Sings}}
		\put(25,17){\textbf{122~\# of Sings}}
		\put(25,-2.0){\textbf{197~\# of Sings}}
		\put(44.8,17){\textbf{175~\# of Sings}}
		\put(44.8,-2.0){\textbf{195~\# of Sings}}
		\put(64.8,17){\textbf{98~\# of Sings}}
		\put(64.8,-2.0){\textbf{160~\# of Sings}}
        \put(84.8,17){\textbf{120~\# of Sings}}
        \put(84.8,-2.0){\textbf{182~\# of Sings}}
	\end{overpic}
 \vspace{2mm}
 \caption{Quad meshes generated by all the methods on two models from ShapeNet~\cite{ShapeNet}~(the airplane model) and Thingi10K~\cite{Thingi10K}~(the grayloc model). We also show the locations of singular points, where~``\# of Sings'' denotes the number of singular points on each quad mesh.}
 \label{fig:sings_and_shape}
\end{figure*}

\paragraph{Fidelity}
In practical applications, when converting a shape from a triangular mesh to a quadrilateral mesh representation, the goals extend beyond minimizing area distortion, angle distortion, and the number of singular points; maintaining fidelity to the original shape is also crucial. Recognizing that a low-resolution quad mesh may naturally lose some details, we use various methods to generate a quad mesh containing 25,000 vertices and 50,000 faces for a more detailed comparison.

In Fig.~\ref{fig:hausdorff}, we present the approximation errors between the quad meshes generated by five methods and the input triangle mesh. The quad meshes generated by our NeurCross and MIQ~\cite{MIQ2009} faithfully represent the original input. IM~\cite{Instant_Meshes2015} and QuadriFlow~\cite{QuadriFlow2018} exhibit minor shape distortions, whereas QuadWild~\cite{quadwild2021} produces a smoother result, leading to a loss of detail.

\paragraph{Resistance to Noise}
As noted in \citet{HessianZX}, \citet{RA2024alignHessian}, and \citet{Dong2024NeurCADRecon}, the Hessian matrix possesses intrinsic smoothing properties. Benefiting from this characteristic, our method demonstrates inherent resistance to noise in cross field prediction. We used a baseline mesh with 15,000 vertices and introduced Gaussian noise (i.e., 2\% relative to the normal direction of each model) to test the noise immunity of our NeurCross. For a comprehensive comparison, we evaluated the four other methods under the same noise conditions.

In Fig.~\ref{fig:noise}, we present the results of different methods under noisy input. Notably, our approach optimizes the SDF and the cross field simultaneously. As a result, during optimization, the underlying SDF naturally smooths out noise, leading to a more intuitive cross field. In summary, our method demonstrates stronger noise resistance compared to four other methods.

\begin{figure*}
    \centering
    \graphicspath{{figures/}}
    \begin{overpic}
		[width=0.98\linewidth]{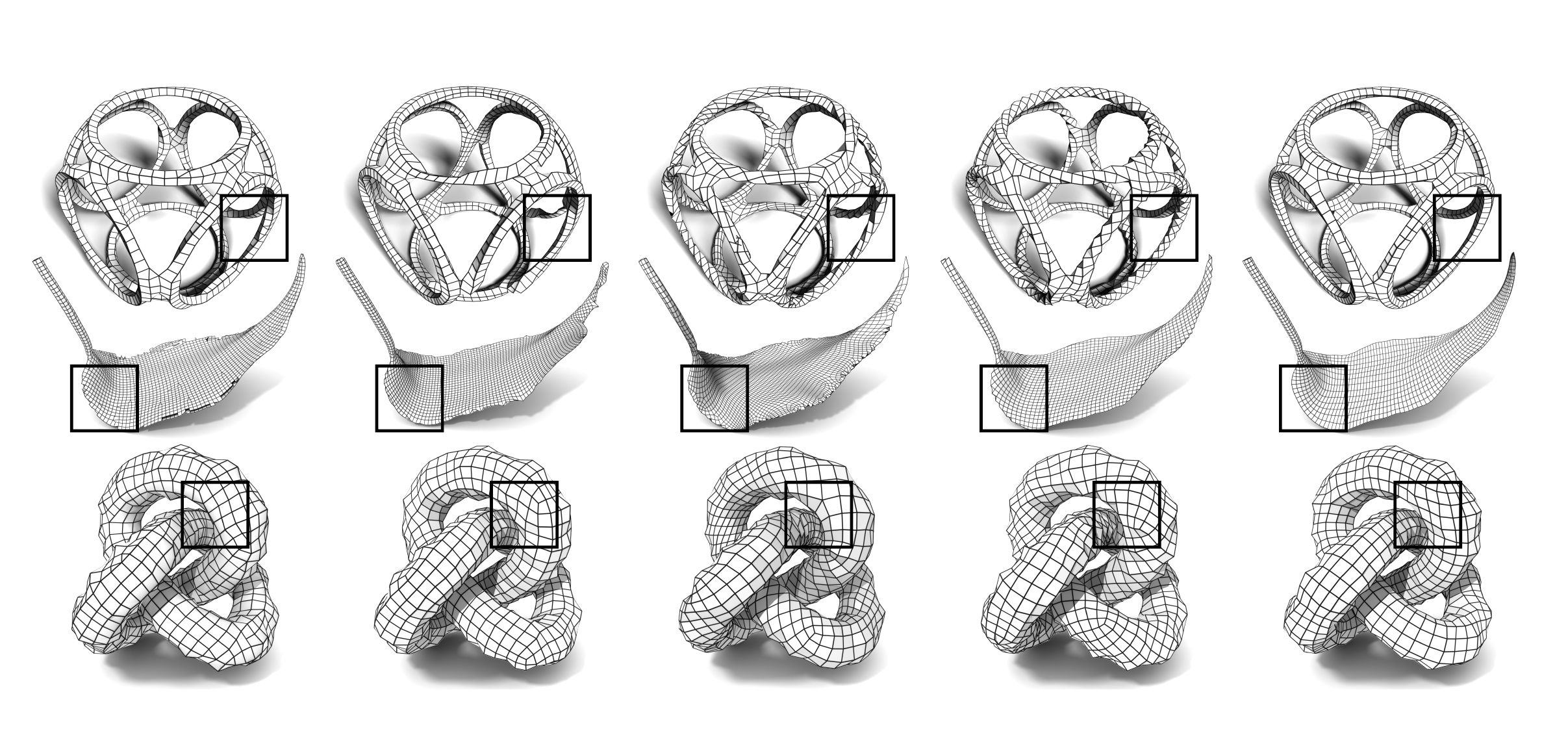}

        \put(8,-3.0){\textbf{IM}}
        \put(25.5,-3.0){\textbf{QuadriFlow}}
        \put(47.2,-3.0){\textbf{QuadWild}}
        \put(69,-3.0){\textbf{MIQ}}
        \put(85,-3.0){\textbf{NeurCross (Ours)}}
    
	\end{overpic}
 \vspace{4mm}
 \caption{
 Comparison of quad meshes generated by various methods for some challenging models, i.e. with high genus, thin shells, and non-orientable rings. Across all tests, the quad meshes generated by  NeurCross consistently exhibit higher quality compared to those produced by other methods.
 }
 \label{fig:complex_model}
 \vspace{-3mm}
\end{figure*}

\begin{figure}[t]
    \centering
    \graphicspath{{figures/}}
    \begin{overpic}
		[width=0.95\linewidth]{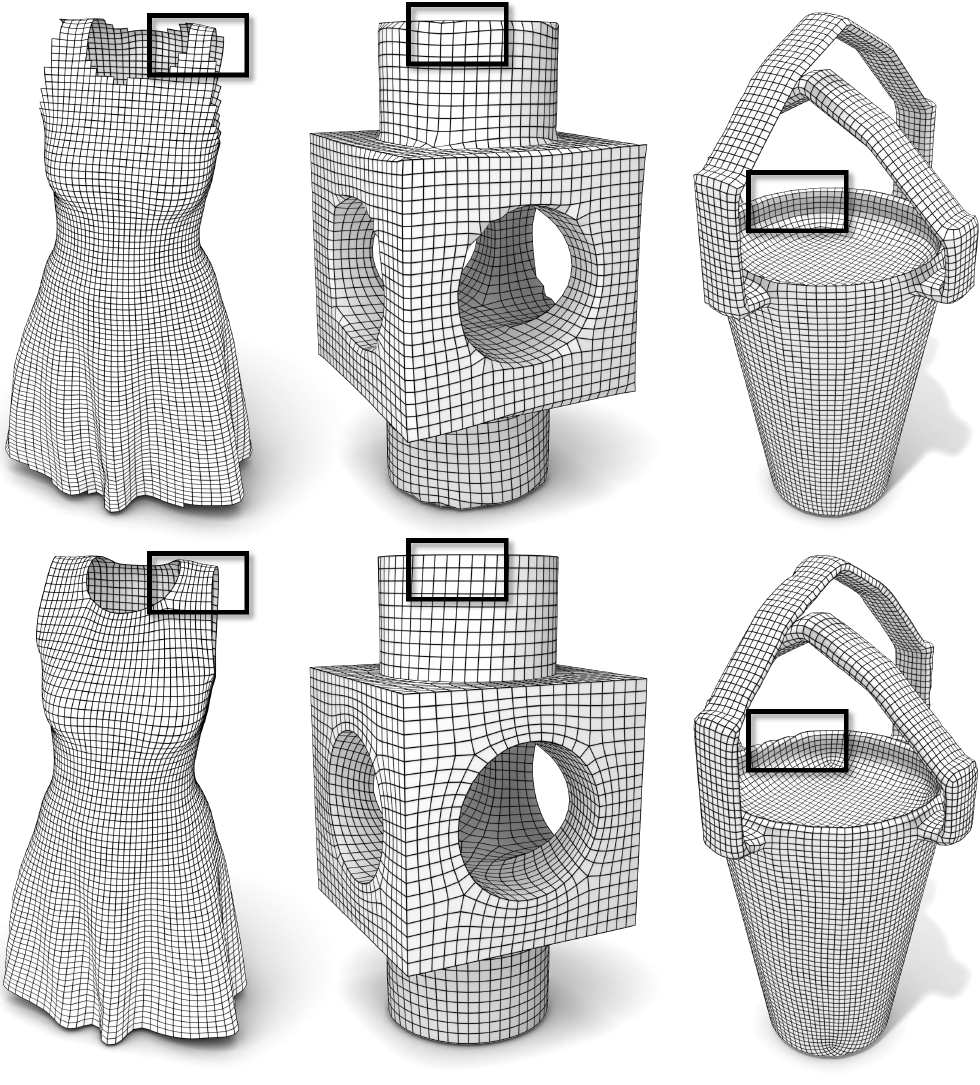}
       \put(-1.5,72){\begin{turn}{90}{\textbf{w/o LPM}} \end{turn}}
       \put(-1.5,22){\begin{turn}{90}{\textbf{w/ LPM}} \end{turn}}
       \put(-1.5,-3.5){\textbf{(a) Open boundaries}}
       \put(32,-3.5){\textbf{(b) Feature lines}}
       \put(60,-3.5){\textbf{(c) Free-form model}}
	\end{overpic}
  \vspace{2.5mm}
 \caption{
 Quad meshes extracted by NeurCross using different mesh extraction methods for various models: (a) A garment with open boundaries; (b) A CAD model with feature lines; and (c) A free-form model. The results are presented for each extraction method with or without the localized patching mechanism~(LPM).
 }
 \label{fig:extraction}
 \vspace{-5mm}
\end{figure}

\begin{figure}[t]
    \centering
    \graphicspath{{figures/}}
    \begin{overpic}
		[width=0.95\linewidth]{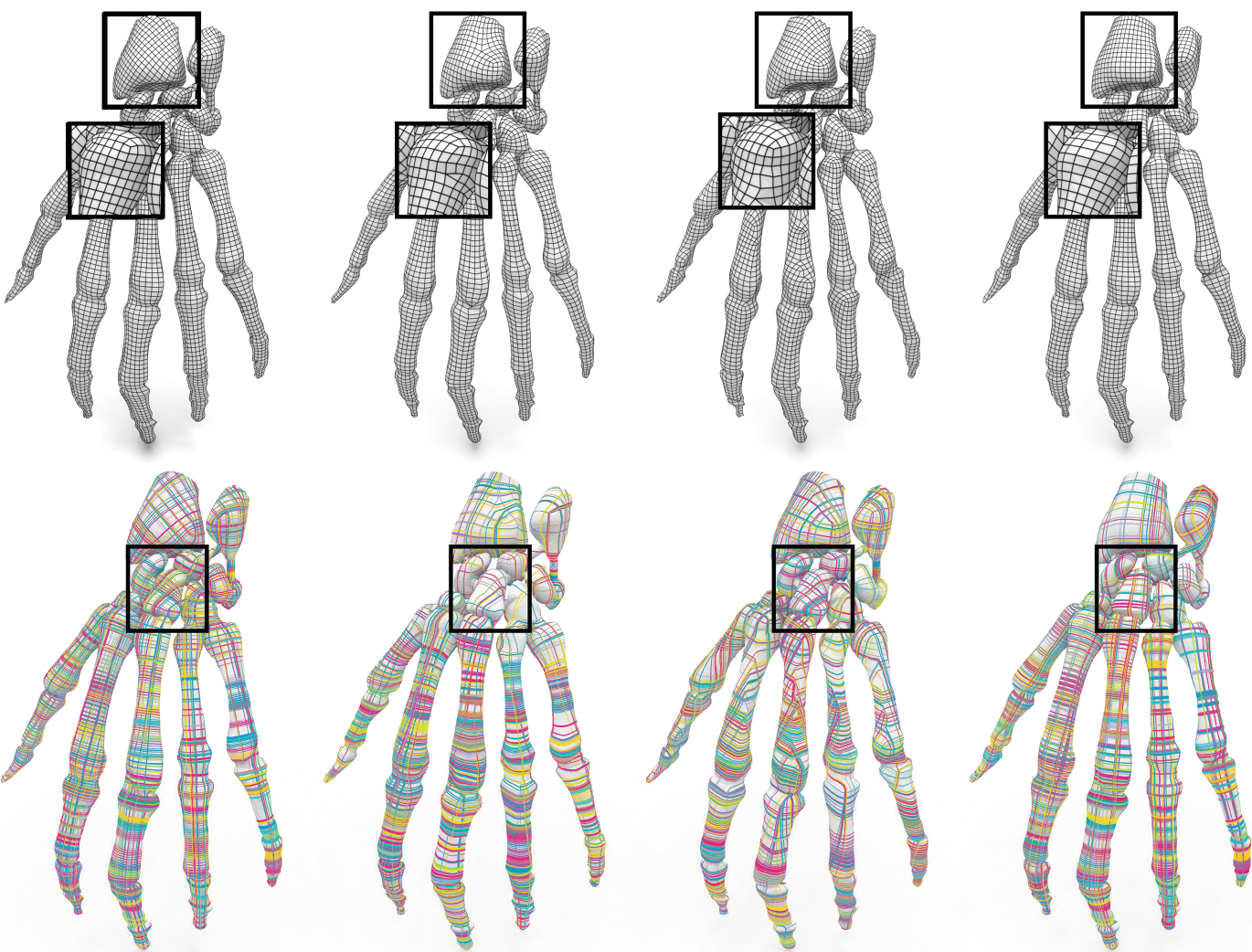}
        \put(3,-6){\textbf{(a) w/o~$\mathcal{L}_\text{AP}$}}
        \put(29,-6){\textbf{(b) w/o~$\mathcal{L}_\text{S}$}}
        \put(50,-6){\textbf{(c) w/o~$\mathcal{L}_\text{AP}$ \&~$\mathcal{L}_\text{S}$}}
        \put(83,-6){\textbf{(d) Ours}}
	\end{overpic}
 \vspace{4mm}
 \caption{
 The quad meshes and cross fields generated by our NeurCross using various loss term combinations: (a) without the alignment with principal directions loss term (w/o $\mathcal{L}_{\text{AP}}$); (b) without the smoothness loss term (w/o $\mathcal{L}_{\text{S}}$);  (c) without both (w/o $\mathcal{L}_{\text{AP}}$ \& $\mathcal{L}_{\text{S}}$); and (d) with both (Ours).
 }
 \label{fig:w_wo_loss_terms}
 \vspace{-4mm}
\end{figure}

\paragraph{Singular Points}
\label{sec:sings}
It's well acknowledged that a trade-off must be achieved between reducing singular points and aligning with principal directions. Thus, it's preferable to position singular points in regions with high curvature variation rather than in flatter areas.
As observed in Tab.~\ref{tab:shapenet}, Tab.~\ref{tab:thingi10k}, and Fig.~\ref{fig:sings_and_shape}, our method produces a slightly higher number of singular points compared to MIQ~\cite{MIQ2009}. This occurrence can be attributed to MIQ's tendency to produce distorted quadrilaterals, which consequently reduces the occurrence of singular points as well as area and angular distortions. However, MIQ's quad mesh lacks overall consistency and tends to oversmooth areas with significant changes in the direction of the cross field.

In Fig.~\ref{fig:sings_and_shape}, we visualize the locations of singular points in the quad meshes generated by all methods on two models. The placement of singular points in the quad mesh generated by our method is more reasonable, and the resulting quadrilateral mesh exhibits high overall consistency.  

\begin{figure*}
    \centering
    \graphicspath{{figures/}}
    \begin{overpic}
		[width=0.89\linewidth]{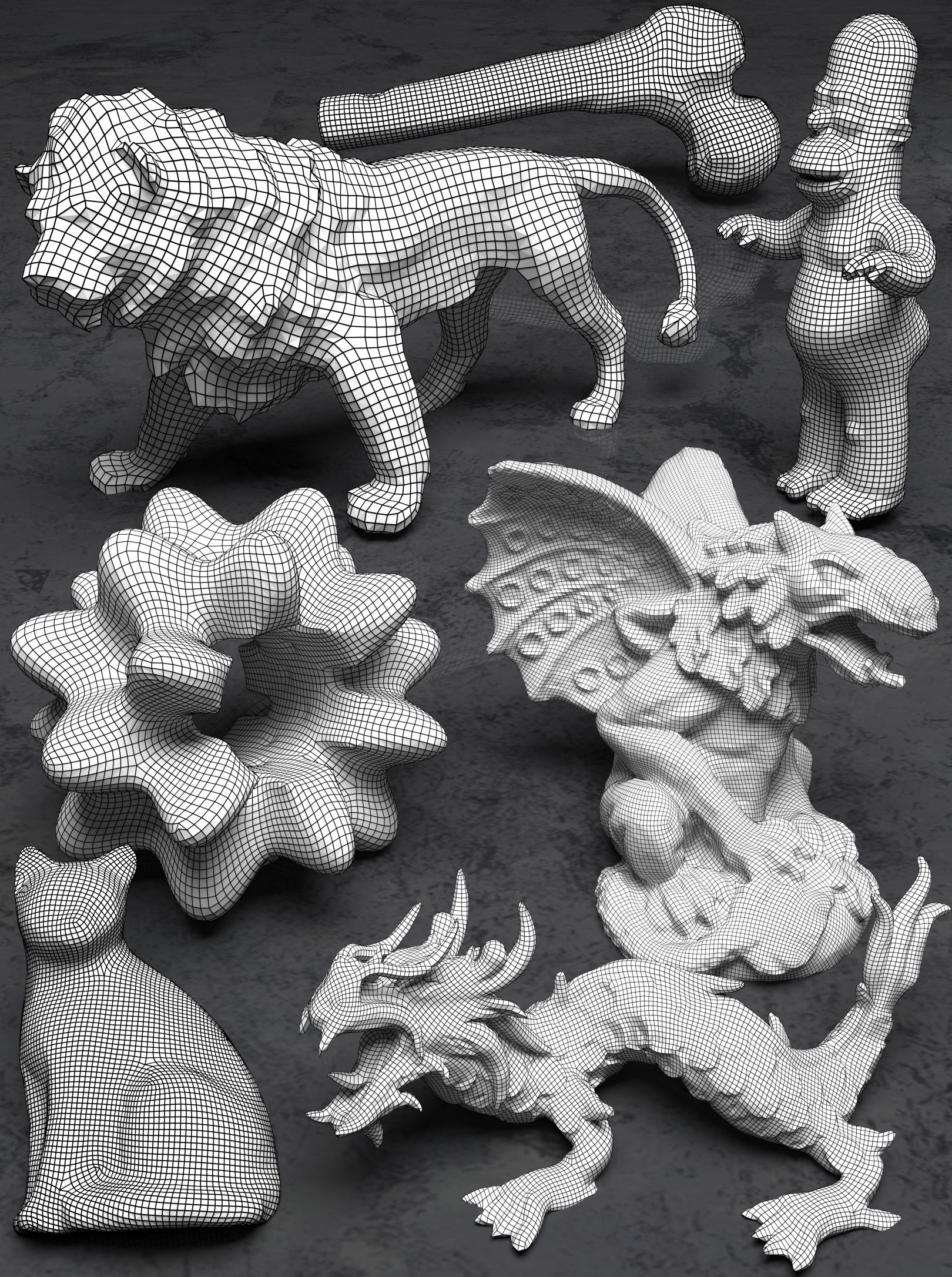}
	\end{overpic}
 \vspace{-2mm}
 \caption{
The quad meshes produced by our NeurCross method on challenging models.
 }
 \label{fig:app_gallery}
\end{figure*}

\paragraph{Geometrically Complex Models}
Various complex geometric models, such as triangular meshes with high genus, thin shells, or non-orientable surfaces, are common in many fields. In Fig.~\ref{fig:complex_model}, we display the quad meshes generated by our method and other methods on several geometrically complex models. The visualization results show that our method's performance on the unoriented ring model is comparable to that of IM~\cite{Instant_Meshes2015} and QuadriFlow~\cite{QuadriFlow2018}. However, for the other two models, only our method consistently produces high-quality quad meshes. Specifically, on the model with a thin shell (the leaf model), only our method and MIQ~\cite{MIQ2009} managed to avoid surface damage. While MIQ produced distorted quadrilaterals at the boundary of the thin shell, our method maintained good overall consistency in the quadrilateral meshes.
Fig.~\ref{fig:app_gallery} shows more results generated by our NeurCros on challenging models.

\begin{figure*}[t]
    \centering
    \graphicspath{{figures/}}
    \begin{overpic}
		[width=0.95\linewidth]{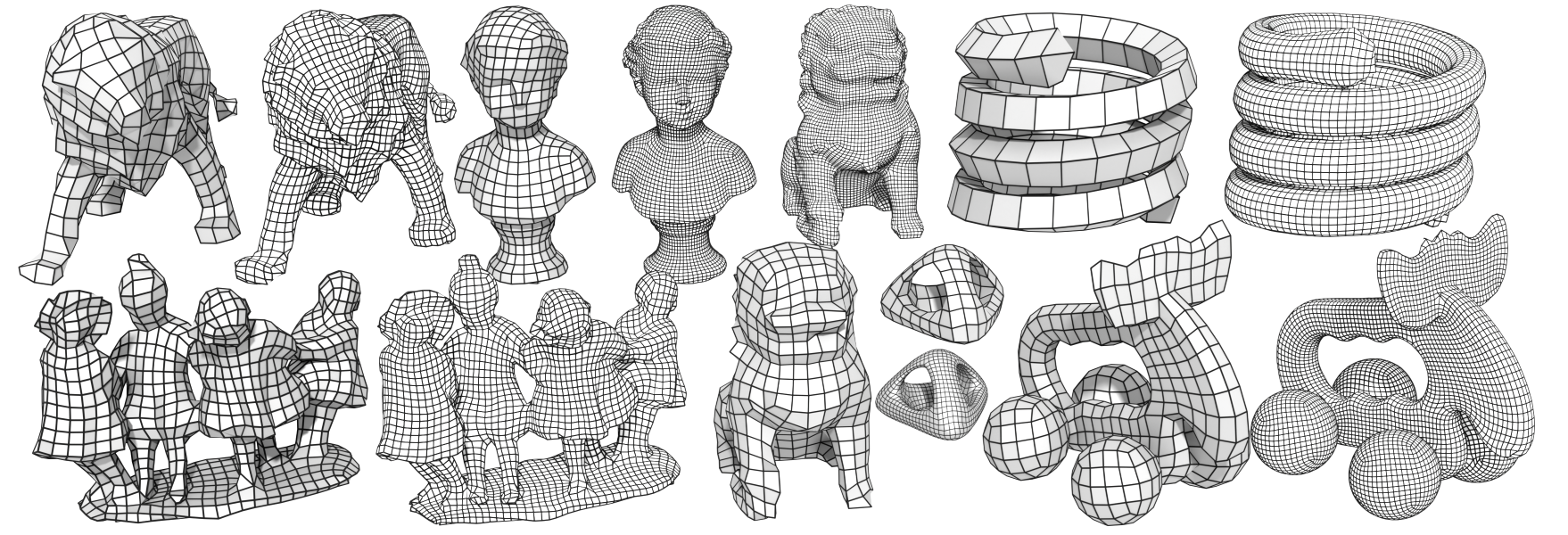}
	\end{overpic}
 \caption{
Quad meshes generated by our NeurCross at different resolutions. The low-resolution models contain fewer than 1,000 vertices, while the high-resolution models consist of over 5,000 vertices.
 }
 \label{fig:resolution}
\end{figure*}

\begin{table}[t]
\vspace{1.5mm}
\centering
\caption{
Ablation studies on the alignment with principal directions loss term $\mathcal{L}_\text{AP}$ and the smoothness loss term $\mathcal{L}_\text{S}$.
}
\label{tab:w_wo_loss_terms}
\resizebox{0.98\linewidth}{!}{
\begin{tabular}{c|c|ccccc} 
\toprule
\multicolumn{2}{c|}{} & Area~$\downarrow$ & Angle~$\downarrow$ & \# of Sings~$\downarrow$ & CD~$\downarrow$ & JR~$\uparrow$\\
\midrule
\multirow{4}{*}{\shortstack{ShapeNet \\~\cite{ShapeNet}}} & w/o~$\mathcal{L}_\text{AP}$ & 1.59 & 11.96 & 89.96 & 8.05 & 0.75\\
~& w/o~$\mathcal{L}_\text{S}$ & 1.96 & 15.12 & 113.28 & 8.09 & 0.71\\
~& w/o~$\mathcal{L}_\text{AP}$ \&~$\mathcal{L}_\text{S}$ & 2.25 & 20.73 & 238.71 & 8.15 & 0.55\\
~& \textbf{NeurCross (Ours)}  & \underline{\textbf{1.48}} & \underline{\textbf{9.85}} & \underline{\textbf{85.32}} & \underline{\textbf{8.03}} & \underline{\textbf{0.78}}\\
\midrule
\multirow{4}{*}{\shortstack{Thingi10K \\~\cite{Thingi10K}}} & w/o~$\mathcal{L}_\text{AP}$ & 1.48 & 11.89 & 73.79 & 8.25 & 0.79\\
~ & w/o~$\mathcal{L}_\text{S}$ & 1.87 & 15.03 & 105.37 & 8.29 & 0.73\\
~& w/o~$\mathcal{L}_\text{AP}$ \&~$\mathcal{L}_\text{S}$ & 2.21 & 20.67 & 225.18 & 8.31 & 0.58\\
~ & \textbf{NeurCross (Ours)} & \underline{\textbf{1.33}} & \underline{\textbf{9.68}}  & \underline{\textbf{68.96}} & \underline{\textbf{8.22}} & \underline{\textbf{0.81}}\\
\bottomrule
\end{tabular}
}
\end{table}

\section{Ablation Studies}

\subsection{Extraction Methods}
As discussed in Section~\ref{sec:Comparison}, the global parameterization techniques in libigl~\cite{libigl2017} fail to align parameterized lines with sharp feature lines. 
To address this, we adopt QuadWild~\cite{quadwild2021}, leveraging the marked sharp features from Sec.~\ref{sec:sharp_feature} to divide the surface into patches using the localized patching mechanism~(LPM)~\cite{quadwild2021}, and using our cross field to guide the patch tessellation process.

As illustrated in the bottom row of Fig.~\ref{fig:extraction}, NeurCross can successfully generate feature-aligned quadrilateral meshes, which is particularly beneficial for CAD models.
For free-form models, the localized patching mechanism~(LPM)~\cite{quadwild2021} introduces singularities at the junctions of adjacent patches and even produces malformed quadrilaterals. 
Therefore, we generally rely on the global parameterization methods from libigl~\cite{libigl2017}, unless the user explicitly requires the alignment of parameterized lines with sharp feature lines, in which case we employ the localized patching mechanism~(LPM)~\cite{quadwild2021}.

\subsection{Cross Field Loss Terms}
To further highlight the efficacy of our cross field loss terms in quad mesh generation, we conducted a comparative analysis by disabling these loss terms. We used the ShapeNet~\cite{ShapeNet} and Thingi10K~\cite{Thingi10K} datasets for testing and comparison, setting the weight $\lambda_{\text{AP}}$ of the alignment with principal directions loss term, the weight $\lambda_{\text{S}}$ of the smoothness loss term, or both, to zero, while keeping other settings unchanged.

Fig.~\ref{fig:w_wo_loss_terms} illustrates the quadrilateral meshes and cross field generated by our method under various loss term combinations. The results show that our method produces the highest quality quadrilateral meshes. Disabling the alignment with principal directions term $\mathcal{L}_{\text{AP}}$ maintains only local correlation and lacks overall consistency. Although the mesh generated without the smoothness term $\mathcal{L}_{\text{S}}$ shows some degree of overall consistency, it is prone to producing singular points due to the absence of constraints on the local cross field. Without constraints from neither $\mathcal{L}_{\text{AP}}$ nor $\mathcal{L}_{\text{S}}$, the resulting quadrilateral mesh exhibits both aforementioned defects. The quantitative results presented in Tab.~\ref{tab:w_wo_loss_terms} align with the qualitative findings in Fig.~\ref{fig:w_wo_loss_terms}, further demonstrating the superiority of our method in generating quadrilateral meshes.

\subsection{Resolution of Quad Mesh}
In real-world applications, selecting the appropriate resolution for quad mesh extraction depends on the specific requirements of different tasks. In Fig.~\ref{fig:resolution}, we use the same cross field for both low- and high-resolution quad meshes, ensuring consistent placement of singular points. Interestingly, the low-resolution mesh better highlights the positioning of these singular points. Fig.~\ref{fig:resolution} demonstrates that, in our approach, most singular points are strategically located in regions with high curvature rather than in flat areas.

\section{limitation}
\begin{figure}[t]
    \centering
    \graphicspath{{figures/}}
    \begin{overpic}
		[width=\linewidth]{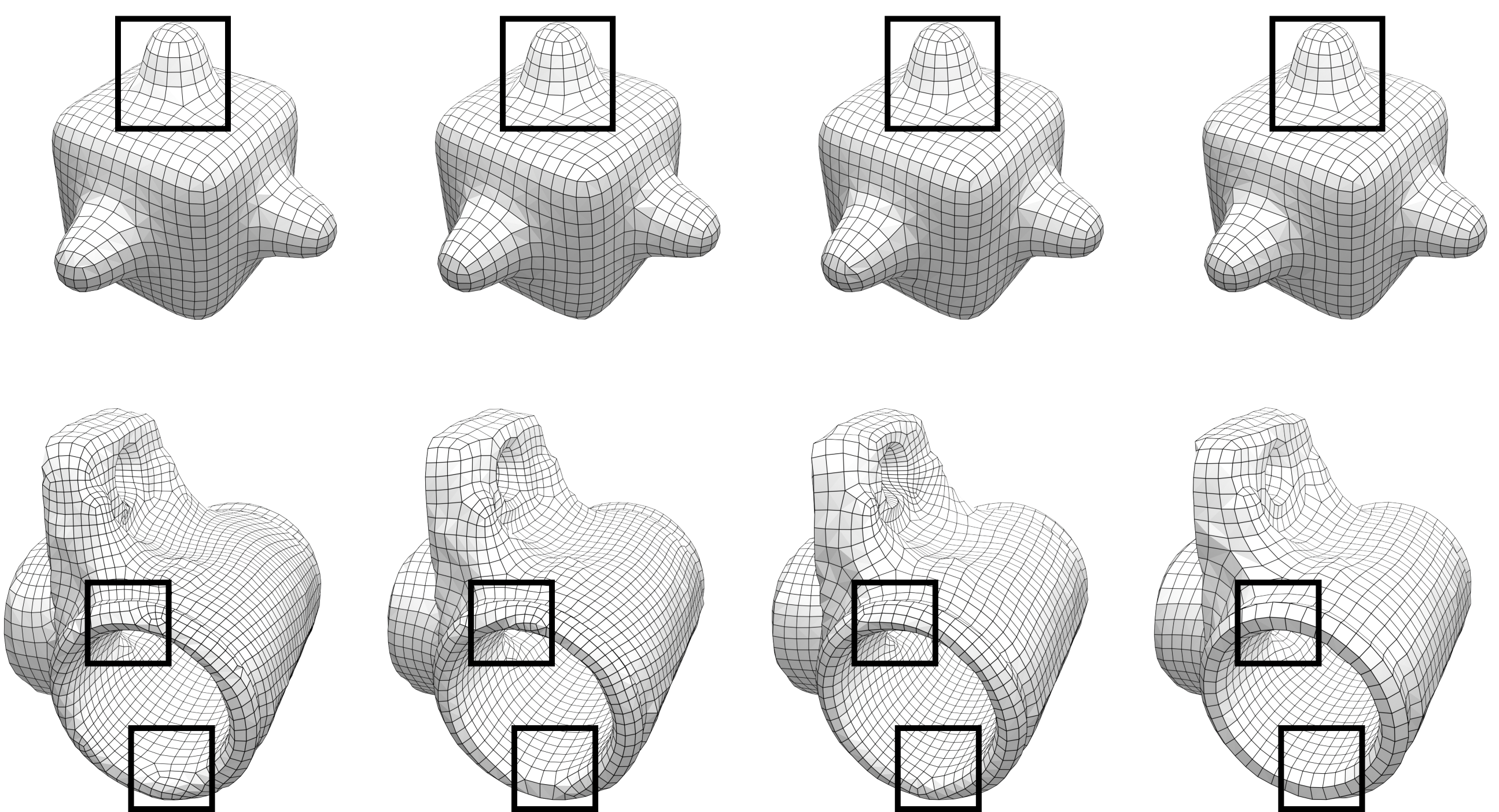}
        \put(4,-6){\textbf{\#iter 500}}
        \put(29,-6){\textbf{\#iter 1k}}
        \put(55,-6){\textbf{\#iter 5k}}
        \put(80,-6){\textbf{\#iter 10k}}
	\end{overpic}
 \vspace{1mm}
 \caption{
Trend of convergence. Quad meshes generated by our NeurCross with different numbers of iterations~(\#iter).
 }
 \label{fig:limitations_times}
\end{figure}

A significant limitation of the self-supervised optimization is its substantial time requirement. For a triangular mesh input with 50,000 faces, each iteration takes 68.34 ms, with a default setting of 10,000 iterations.
However, for geometrically simple and regular shapes, NeurCross typically converges in fewer iterations to produce high-quality quadrilateral meshes~(see the top row of Fig.~\ref{fig:limitations_times}), whereas complex shapes may require additional iterations to achieve comparable results~(see the bottom row of Fig.~\ref{fig:limitations_times}).

A promising future direction is to leverage this approach to generate ample training data for feeding generative models, such as MeshGPT~\cite{Meshgpt2024}. This would enable users to obtain high-quality quad meshing outcomes instantly.

\section{conclusion}
In this paper, we propose a self-supervised neural representation of the cross field for quadrilateral mesh generation. To the best of our knowledge, this is the first self-supervised approach for this task. Our network, named NeurCross, consists of two modules: one to fit the SDF and another to predict the cross field. The design of our loss function addresses three key aspects: surface approximation quality, alignment with principal directions, and the spatial smoothness of the cross field. Leveraging our network, the SDF and cross field are optimized simultaneously, achieving a desirable balance between approximation accuracy and cross field smoothness. Experimental results consistently validate improvements in singular point placement and in the approximation accuracy between the input triangular surface and the output quad mesh.

\section*{Acknowledgments}
The authors would like to thank the anonymous reviewers for their valuable comments and suggestions. This work was supported by the National Key R\&D Program of China (2022YFB3303200), the National Natural Science Foundation of China (U23A20312, 62272277, 62102380), the Shandong Provincial Natural Science Foundation (ZR2024MF083), the Innovation and Technology Commission of the HKSAR Government under the InnoHK initiative (TransGP project) and the ITSP-Platform grant (Ref: ITS/335/23FP), and the Research Grants Council of Hong Kong (Ref: 17210222).

\bibliographystyle{ACM-Reference-Format}
\bibliography{main}

\clearpage

\end{document}